%% file: aaai2026.tex
\documentclass[letterpaper]{article} 
\usepackage{aaai2026}  
\usepackage{times}  
\usepackage{helvet}  
\usepackage{courier}  
\usepackage[hyphens]{url}  
\usepackage{graphicx} 
\usepackage{natbib}  
\usepackage{caption} 
\usepackage{algorithm}
\usepackage{algorithmic}
\usepackage{amsmath}
\usepackage{amsfonts}
\usepackage{makecell}
\usepackage{booktabs}
\usepackage{multirow}
\usepackage{xcolor}
\usepackage[table]{xcolor}

\usepackage{newfloat}
\usepackage{listings}
\DeclareCaptionStyle{ruled}{labelfont=normalfont,labelsep=colon,strut=off} 
\floatstyle{ruled}
\newfloat{listing}{tb}{lst}{}
\floatname{listing}{Listing}
\title{SplatSSC: Decoupled Depth-Guided Gaussian Splatting for Semantic Scene Completion}
\author{ 
    Rui Qian\equalcontrib\textsuperscript{\rm 1}, 
    Haozhi Cao\equalcontrib\textsuperscript{\rm 1}, 
    Tianchen Deng\textsuperscript{\rm 2},
    Shenghai Yuan\textsuperscript{\rm 1},
    Lihua Xie\textsuperscript{\rm 1}
} 
\affiliations{
    \textsuperscript{\rm 1}Nanyang Technological University\\
    \textsuperscript{\rm 2}Shanghai Jiao Tong University\\
    \{rqian003, haozhi002, shyuan, elhxie\}@ntu.edu.sg, dengtiancheng@sjtu.edu.cn
}

\begin{document}

\maketitle 

\begin{abstract}
Monocular 3D Semantic Scene Completion (SSC) is a challenging yet promising task that aims to infer dense geometric and semantic descriptions of a scene from a single image. While recent object-centric paradigms significantly improve efficiency by leveraging flexible 3D Gaussian primitives, they still rely heavily on a large number of randomly initialized primitives, which inevitably leads to 1) inefficient primitive initialization and 2) outlier primitives that introduce erroneous artifacts. In this paper, we propose SplatSSC, a novel framework that resolves these limitations with a depth-guided initialization strategy and a principled Gaussian aggregator. Instead of random initialization, SplatSSC utilizes a dedicated depth branch composed of a Group-wise Multi-scale Fusion (GMF) module, which integrates multi-scale image and depth features to generate a sparse yet representative set of initial Gaussian primitives. To mitigate noise from outlier primitives, we develop the Decoupled Gaussian Aggregator (DGA), which enhances robustness by decomposing geometric and semantic predictions during the Gaussian-to-voxel splatting process. Complemented with a specialized Probability Scale Loss, our method achieves state-of-the-art performance on the Occ-ScanNet dataset, outperforming prior approaches by over 6.3\% in IoU and 4.1\% in mIoU, while reducing both latency and memory cost by more than 9.3\%. 
\end{abstract}

\begin{links}
\link{Code}{https://github.com/Made-Gpt/SplatSSC}
\end{links}

\begin{figure}[t]
\centering
\includegraphics[width=0.95\columnwidth]{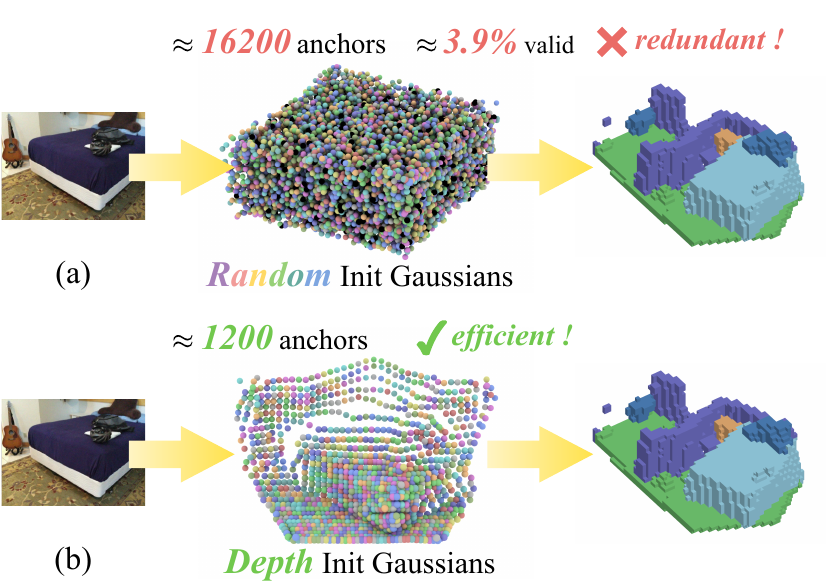}
\caption{
Comparison with prior framework. (a) Recent transformer-based SSC frameworks start with a large set of randomly initialized Gaussian primitives, introducing redundancy. (b) Our framework starts with a compact yet targeted set of Gaussian primitives, guided by geometric priors. }
\label{fig1}
\end{figure}

\section{Introduction} 

3D scene understanding has garnered significant attention with the rapid evolution of embodied agents and autonomous driving~\cite{li2025flagdroneacing, yi2025embodiedgame, fusic2024improvedrrt, zheng2025robustefficient, zammit2023realtimeuav}. As a key technology in this domain, 3D occupancy prediction~\cite{tong2023scene, huang2023tri, wei2023surroundocc, tian2023occ3d, wang2024panoocc} and 3D Semantic Scene Completion (SSC)~\cite{cao2024monoscene, miao2023occdepth, zhang2023occformer, li2023voxformer, mei2024camera} have made remarkable progress. Early and conventional approaches for these tasks predominantly rely on grid-based representations. However, processing dense 3D volumes incurs prohibitive computational and memory costs. To mitigate this limitation, various efficiency-driven strategies have been explored, such as accelerating processing with Bird's-Eye-View (BEV) projections~\cite{yu2023flashocc, hou2024fastocc}, or leveraging the natural sparsity of scenes with sparse voxels~\cite{tang2024sparseocc, li2023voxformer} and points~\cite{shi2024occupancy, wang2024opus}.
 
Despite the devoted efforts, such methods remain inherently constrained by their discrete and grid-aligned nature, which struggle to model the sparse geometry efficiently. A recent paradigm shift towards object-centric representations, pioneered by GaussianFormer~\cite{huang2024gaussianformer}, has achieved a breakthrough. By utilizing flexible 3D Gaussian primitives~\cite{kerbl20233d} to represent the scene, this approach strikes a new balance between performance and efficiency. Building upon this foundation, subsequent works~\cite{huang2025gaussianformer2} have advanced this field by developing more principled aggregation methods based on Gaussian Mixture Models (GMMs) and adapting the paradigm to indoor scenes for incremental perception~\cite{wu2024embodiedocc, zhang2025roboocc, wang2025embodiedoccplusplus}.

While the object-centric paradigm offers a promising direction, its application in vision-only settings faces a foundational challenge: \textit{\textbf{how to efficiently initialize and reliably supervise 3D primitives using only monocular cues}}. To ensure complete coverage of the target 3D space without geometric cues, the predominant strategy is to randomly distribute numerous primitives throughout the 3D volume, as shown in Figure~\ref{fig1}(a). This leads to two critical, coupled limitations: 
1) \textit{Inefficient Primitive Initialization.} A significant portion of the model's capacity is inevitably wasted on representing empty or unknown space, making the random distribution strategy inherently redundant. 
2) \textit{Fragile Aggregation of Outliers.} Existing Gaussian-to-voxel splatting strategies~\cite{huang2024gaussianformer,huang2025gaussianformer2} lack an effective rejection mechanism to mitigate the impact of outlier primitives. This allows outliers to spurious semantics on distant voxels, creating ``floaters'' in otherwise empty space. 

To this end, we introduce SplatSSC, a novel framework designed to tackle inefficient initialization and fragile aggregation in object-centric SSC. 
Rather than starting with a large set of random primitives, SplatSSC leverages geometric priors to guide the primitive initialization, as shown in Figure~\ref{fig1}(b), reducing redundancy while maintaining representational capacity.
We generate this prior using a tailored depth branch, equipped with our proposed \textit{Group-wise Multi-scale Fusion} (GMF) module. 
GMF integrates multi-scale image features and depth features from a pretrained depth estimator via \textit{Group Cross-Attention} (GCA) for efficient multi-modal fusion. 
The resulting geometric priors subsequently guide a lifter to initialize a sparse yet targeted set of Gaussian primitives, which are then refined through a standard multi-stage encoder. 
To address the ``floaters'' that plague existing aggregators when dealing with sparse outliers, we propose the \textit{Decoupled Gaussian Aggregator} (DGA), which renders the final semantic grid by completely decomposing semantic and geometry prediction robustly. 
Furthermore, to ensure stable geometric learning, we design the specialized \textit{Probability Scale Loss} to apply soft and progressive supervision to the intermediate encoder layers.

In summary, our contributions are as follows:
\begin{itemize}
    \item We propose an efficient object-centric paradigm for monocular SSC, namely SplatSSC, which features a depth-guided strategy for initializing a sparse and targeted set of Gaussian primitives.
    \item We introduce the \textit{Group-wise Multi-scale Fusion (GMF)} module with a \textit{Group Cross-Attention (GCA)} core to efficiently generate a high-quality geometric prior.
    \item We design the \textit{Decoupled Gaussian Aggregator (DGA)} that decouples geometry and semantics to eliminate aggregation artifacts from sparse primitives robustly.
    \item We propose a \textit{Probability Scale Loss} to provide auxiliary geometric supervision for robust end-to-end training.
\end{itemize}

\section{Related Work}

\subsubsection{3D Semantic Scene Completion.}

3D Semantic Scene Completion (SSC) infers dense geometry and semantics from partial observations. Early approaches~\cite{song2017sscnet, zhang2019cascaded, wang2019forknet} primarily focused on indoor scenes using depth-only input, where deep convolutional networks (CNNs) and Truncated Signed Distance Function (TSDF) representations were widely employed. To improve semantic understanding, subsequent methods~\cite{li2019rgbd, li2020anisotropic, wang2023semantic} fuse features from both RGB and depth inputs. In parallel, LiDAR-based SSC approaches~\cite{roldao2020lmscnet, yan2021sparse, yang2021semantic} have been developed for autonomous driving and also rely on CNN architectures.

A recent trend has shifted towards vision-only methods. MonoScene~\cite{cao2024monoscene} pioneered this direction using a dense 2D-to-3D lifting with UNet architecture~\cite{ronneberger2015u}, but this approach suffered from inherent depth ambiguity. To address this, OccDepth~\cite{miao2023occdepth} and ISO~\cite{yu2024monocular} introduced depth-aware strategies by leveraging stereo depth and pretrained depth networks, respectively. Concurrently, to tackle the inefficiency of dense voxel processing, VoxFormer~\cite{li2023voxformer}, a two-stage model, proposed a sparse-to-dense Transformer method based on generating proposals from a geometry prior. Subsequent works continue to advance this paradigm~\cite{mei2024camera, zhu2024CGFormer, jiang2024symphonize}, focusing on unified pipelines, context-aware modeling, and instance-level reasoning.

While these Transformers improve accuracy, they remain bound to grid-aligned voxel queries. Our work takes a different route through a flexible object-centric formulation inspired by~\cite{wu2024embodiedocc, huang2025gaussianformer2}.

\begin{figure*}[t]
\centering
\includegraphics[width=0.95\textwidth]{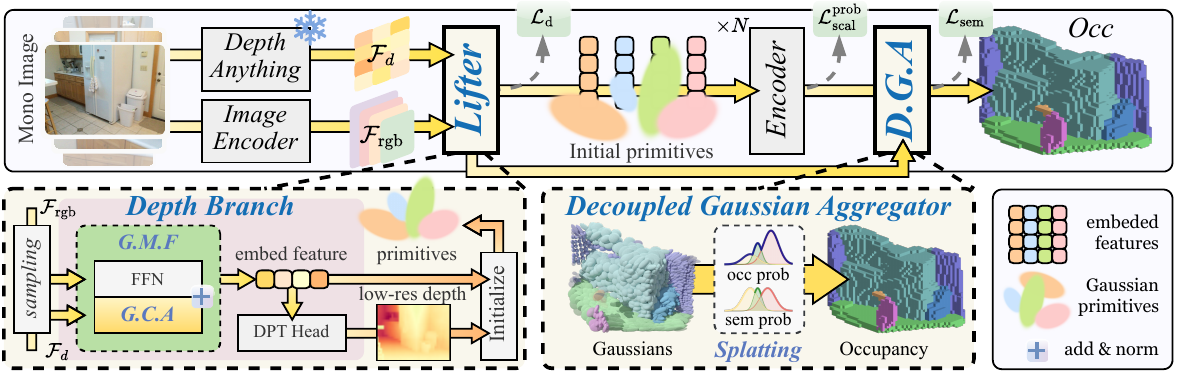} %
\caption{An overview of our proposed SplatSSC architecture. Given a single input image, our model employs two parallel branches: a trainable image encoder to extract multi-scale image features, and a frozen, pretrained \textit{Depth-Anything }model to extract depth features. After a sampling step, both features are fed into the proposed \textit{Group-wise Multi-scale Fusion} (GMF) block and a two-convolution layer depth head, yielding a refined feature map and a low-resolution depth map. These outputs are then lifted to initialize a set of 3D Gaussian primitives. Subsequently, the primitives are processed by a multi-stage encoder and finally passed to our \textit{Decoupled Gaussian Aggregator} (DGA) to render the final semantic voxels.}
\label{fig2}
\end{figure*}

\subsubsection{Object-centric 3D Scene Representation.}
A recent paradigm shift in occupancy prediction, pioneered by GaussianFormer~\cite{huang2024gaussianformer}, moves beyond grid-aligned queries to object-centric representation using 3D Gaussian primitives. This approach leverages the inherent sparsity of 3D scenes by representing them as a collection of continuous ellipsoids, which are then rendered into a dense semantic grid via an efficient Gaussian-to-voxel splatting mechanism. This marked a significant departure from discrete, voxel-based frameworks~\cite{li2023voxformer, tang2024sparseocc}.
 
Sequential works~\cite{huang2025gaussianformer2, zhao2025gaussianformer3d} further advanced this paradigm by introducing a principled probabilistic framework via GMMs and incorporating LiDAR-guided initialization to replace random placement. In parallel, EmbodiedOcc~\cite{wu2024embodiedocc} first adapted this object-centric paradigm to the unique challenges of indoor perception. It focuses on online incremental scene understanding, where confidence refinement is applied to continuously update the Gaussian representation as an agent explores the environment. Following this, RoboOcc and EmbodiedOcc++~\cite{zhang2025roboocc, wang2025embodiedoccplusplus} extended this paradigm through geometry-aware refinement, leveraging opacity cues and planar constraints to enhance stability and structural fidelity.

However, object-centric approaches widely employ random Gaussian primitive initialization, which introduces significant redundancy, as most primitives are used to represent empty space. In contrast, our method directly tackles this problem by leveraging a depth prior to generate a compact but more targeted set of primitives. 

\section{Methodology}

\subsection{Problem Setup}
Formally, given a single input RGB image $\mathcal{I}_\mathrm{rgb}$, the local prediction task is to infer the dense semantic voxel grid $\mathcal{V}_{\mathrm{loc}}$ and the underlying set of sparse Gaussian primitives $\mathcal{G}_\mathrm{loc}$ that represent the scene within the current camera frustum. This process is defined as:
\begin{equation}
    (\mathcal{V}_{\mathrm{loc}}, \mathcal{G}_\mathrm{loc}) = \mathcal{M}_{\mathrm{loc}}(\mathcal{I}_\mathrm{rgb}), 
\end{equation}
where \begin{small}$\mathcal{M}_{\mathrm{loc}}$\end{small} is our prediction model. The output grid $\mathcal{V}_{\mathrm{loc}} \in \{0, 1, ..., C-1\}^{X_\mathrm{loc} \times Y_\mathrm{loc} \times Z_\mathrm{loc}}$ assigns each voxel a label from $C$ semantic classes, with class 0 denoting empty space. The scene itself is represented by the set of $N$ refined Gaussian primitives $\mathcal{G}_\mathrm{loc} = \{G_i\}_{i=1}^N$. Each primitive $G_i$ is parameterized by its geometric and semantic properties: a mean $\boldsymbol{\mu}_i \in \mathbb{R}^3$, a scale vector $\mathbf{s}_i \in \mathbb{R}^3$, a rotation quaternion $\mathbf{q}_i \in \mathbb{R}^4$, an opacity $\mathbf{a}_i \in [0, 1]$, and a semantic logit vector $\mathbf{c}_i \in \mathbb{R}^{C-1}$. The scale and rotation are used to construct the full anisotropic covariance matrix $\mathbf{\Sigma}_i$:
\begin{equation} \label{eq:covariance} 
    \mathbf{\Sigma}_i = \mathbf{R}_i \mathbf{S}_i \mathbf{S}_i^T \mathbf{R}_i^T, \, \mathbf{S}_i = \mathrm{diag}(\mathbf{s}_i), \, \mathbf{R}_i = \mathrm{q2r}(\mathbf{q}_i).
\end{equation}
where $\mathrm{q2r}(\cdot)$ converts a quaternion into a rotation matrix and $\mathrm{diag}(\cdot)$ forms a diagonal scaling matrix.

\subsection{Overview}
The architecture of our approach is illustrated in Figure~\ref{fig2}. We first process the input image $\mathcal{I}_{\mathrm{rgb}}$ with an image encoder, composed of a lightweight image backbone EfficientNet~\cite{tan2019efficientnet} and FPN~\cite{lin2017feature}, to extract multi-scale image features $\mathcal{F}_{\mathrm{rgb}}=\{f_{\mathrm{rgb}}^{l}\}_{l=1}^L$, where $L$ is the scale number. Simultaneously, a pretrained depth estimation model \textit{Depth-Anything}~\cite{yang2024depth} is employed to produce powerful depth features $\mathcal{F}_d$. These two feature streams are then fed into our specialized \textit{depth branch}, which employs the proposed GMF module to produce the fused depth features $\mathcal{F}_{d}'$ and the refined depth map $\mathcal{I}_{d}$. The resulting $\mathcal{F}_{d}'$ and $\mathcal{I}_d$ are then fed to a lifting module to obtain the initial Gaussian primitives $\mathcal{G}_o$ with good geometry prior. Subsequently, $\mathcal{G}_{o}$ is refined by a series of encoder blocks cyclically, following EmbodiedOcc. Given the refined primitives, the 3D semantic voxels are obtained by our DGA \begin{small}$\hat{\mathcal{V}}_{\mathrm{agg}}$\end{small}. By first leveraging the depth branch to generate a highly compact set of primitives with geometrically grounded initial locations, we tackle the inefficiency inherent in random initialization strategies. Subsequently, our DGA transforms primitives into semantic voxels, overcoming the fragility of prior aggregation methods. This enables our framework to achieve state-of-the-art (SOTA) performance while maintaining high efficiency with significantly fewer primitives.

\subsection{Depth Branch}
While recent monocular 3D completion methods~\cite{wu2024embodiedocc, yu2024monocular} leverage pretrained depth estimators, they tend to utilize depth information as a secondary guiding signal: either refining geometric distributions or informing feature learning. However, this approach neglects the rich latent features generated by depth networks. In contrast, our framework proposes a dual-pronged strategy: we use the depth map as a direct geometric prior, while simultaneously employing the latent depth features as the initial embeddings for 3D primitives. This not only ensures primitives are grounded in both geometry (\textit{where}) and semantics (\textit{what}), but necessitates a more advanced fusion mechanism. To fulfill this demand, we design a dedicated depth branch. Inspired by prior works~\cite{ma2020auto, jia2025gated}, this branch fuses multi-scale image features and depth cues via our GMF mechanism. Specifically, GMF is a Transformer-like block comprising the proposed GCA layer followed by a point-wise FFN~\cite{ashish2017transformer}. The resulting fused features $\mathcal{F}_{d}'$ are then processed by two convolutional layers to produce the refined depth map $\mathcal{I}_d$. 

\begin{figure}[t]
\centering
\includegraphics[width=0.93\columnwidth]{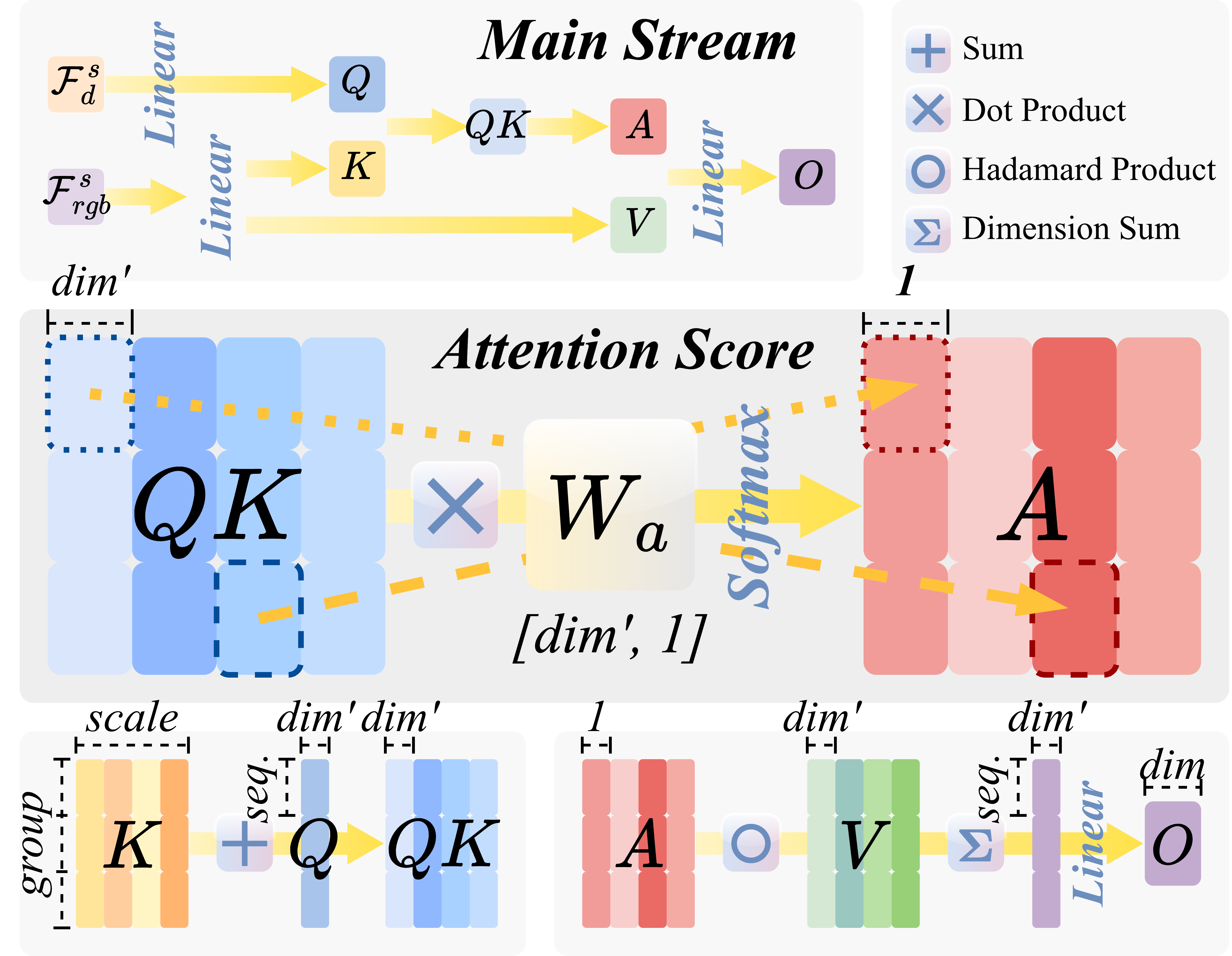} 
\caption{Illustration of the proposed GCA layer. The weight matrix $W_a$ is shared across different groups and scales, thus reducing memory consumption and computational cost.} 
\label{fig3} 
\vspace{-13pt}
\end{figure} 

\subsubsection{Group Cross-Attention.}
The architecture of our GCA module is illustrated in Figure~\ref{fig3}.
The process begins by sampling features from the input depth features $\mathcal{F}_d$ and multi-scale image features $\mathcal{F}_{\mathrm{rgb}}$, using a set of predefined reference points normalized to the $[0,1]$ range. This step yields the sampled features, denoted as $\mathcal{F}_d^s$ and \begin{small}$\mathcal{F}_{\mathrm{rgb}}^s = \{f_{\mathrm{rgb}}^{s,l}\}_{l=1}^L$\end{small} respectively.
To balance performance and efficiency, we split these features into $G$ groups along the channel dimension, where each group has a reduced feature dimensionality of \begin{small}$D_g = D / G$\end{small}. The Query $Q_g$ is projected from sampled depth features, while the Key \begin{small}$K_g^l$\end{small} and Value \begin{small}$V_g^l$\end{small} are projected from sampled image features at each scale $l$:
\vspace{-3pt}
\begin{equation} 
    Q_g = ({\mathcal{F}_d^{s} W_q})^g,
    K_g^l = ({f_{\mathrm{rgb}}^{s,l} W_k})^g,
    V_g^l = ({f_{\mathrm{rgb}}^{s,l} W_v})^g,
\end{equation}
where $W_q$, $W_k$, and $W_v$ are linear projection matrices for Query, Key, and Value, respectively. $l \in \{1, ..., L\}$ denotes the scale index. Inspired by the efficient design of Deformable Attention~\cite{zhu2020deformable}, we adopt a lightweight linear projection mechanism in place of the standard dot-product attention. To elaborate, the attention scores are computed by feeding the element-wise sum of queries and keys into a shared projection \begin{small}$W_a \in \mathbb{R}^{D_g \times 1}$\end{small}:
\begin{equation}
    A_g^l = \mathbb{S}_l \left( W_a (Q_g + K_g^l) \right),
\end{equation}
where $\mathbb{S}_l(\cdot)$ denotes the Softmax operation across the scale dimension, and $g$ indexes feature groups. With the group-wise formulation, both scale-wise attention and projection are computed within each group, allowing $W_a$ to be shared across different groups and scales. This design significantly reduces parameter overhead and computation.

The final fused representation is obtained by aggregating value features $V_g^l$ using Hadamard product $\circ$ with the attention scores, followed by group concatenation \begin{small}$\mathbb{C}_g(\cdot)$\end{small} and a linear projection $W_o$:
\vspace{-3pt}
\begin{equation}
    \mathcal{F}_{d}' = \mathbb{C}_{g} \left( \sum_{l=1}^L A_g^l \circ V_g^l \right) W_o.
\end{equation}

\subsubsection{Efficiency Analysis.}
The design of GCA is computationally lean. Standard cross-attention has a complexity of \begin{small}$\mathcal{O}(LN^2D)$\end{small}, where $N$ is the sequence length. In contrast, by employing a group-wise mechanism and replacing the quadratic-cost dot-product with a linear-cost MLP, GCA significantly reduces the complexity. The dominant cost of our module becomes \begin{small}$\mathcal{O}(ND^2(L+2) / G)$\end{small}, which is substantially more efficient, especially for long sequences.

\subsection{Decoupled Gaussian Aggregator}\label{subsec:DGA}

Gaussian-to-voxel splatting is a critical step for object-centric approaches, which dictates the final quality of the occupancy output. While GaussianFormer first enabled object-centric aggregation, its additive nature leads to redundancy. The subsequent \textit{Probabilistic Gaussian Superposition} (PGS) model proposed in GaussianFormer-2, though theoretically elegant, introduces a flawed decoupling of geometry and semantics and therefore falls short when tackling outlier primitives. To address these limitations, we propose the DGA, a novel strategy that reformulates the task into two distinct prediction pathways: \textit{Geometry Occupancy Prediction} and \textit{Conditional Semantic Distribution}. 

\subsubsection{Analysis of Probabilistic Gaussian Superposition.}
The PGS models the semantic occupancy prediction at a point $\mathbf{x}$ as a two-part process: a geometric occupancy probability $\alpha(\mathbf{x})$ and a conditional semantic expectation $\mathbf{e}(\mathbf{x};\mathcal{G})$:
\begin{equation}
    \alpha(x) = 1 - \prod_{i \in \mathcal{N}(\mathbf{x})} \left(1 - \alpha(\mathbf{x};G_i)\right),
\end{equation}
\begin{equation}
    \mathbf{e}(\mathbf{x}; \mathcal{G}) = \sum_{i=1}^N p(G_i|\mathbf{x}) \tilde{\mathbf{c}}_i = \frac{\sum_{i=1}^N p(\mathbf{x}|G_i) \mathbf{a}_i \tilde{\mathbf{c}}_i}{\sum_{j=1}^N p(\mathbf{x}|G_j) \mathbf{a}_j},
\end{equation} 
\begin{equation}
    p(\mathbf{x}|G_i) = \frac{1}{(2\pi)^{3/2} |\mathbf{\Sigma}_i|^{1/2}} \alpha(\mathbf{x}; G_i),
\end{equation}
where $p(\mathbf{x}\mid G_i)$ denotes the conditional Gaussian probability of point $\mathbf{x}$ under the $i$th primitive, and $p(G_i\mid \mathbf{x})$ represents the normalized posterior of selecting $G_i$ for $\mathbf{x}$ under a uniform prior.
$\alpha(\mathbf{x}; G_i) = \exp(-\frac{1}{2}(\mathbf{x}-\boldsymbol{\mu}_i)^T \mathbf{\Sigma}_i^{-1}(\mathbf{x}-\boldsymbol{\mu}_i))$ is the un-normalized Gaussian kernel. 
The key flaw in this formulation lies in how the learned opacity $\mathbf{a}_i$ is used. While intended to represent a primitive's existence confidence, it is instead employed as the prior probability in the GMM. The negative consequence of this choice becomes evident when considering an isolated outlier primitive $G_n$. For any point $\mathbf{x}^f$ in its immediate vicinity, the likelihood $p(\mathbf{x}^f|G_m)$ for all other distant primitives $G_{m \neq n}$ approaches zero. This causes the normalization term in the posterior calculation to be dominated by the outlier itself. Hence, the posterior probability $p(G_n|\mathbf{x}^f)$ collapses to unity, regardless of the effect of the low-confidence prior $\mathbf{a}_n$:
\begin{align}
    p(G_n|\mathbf{x}^f) & = \frac{p(\mathbf{x}^f|G_n) \mathbf{a}_n}{\sum_{j=1}^N p(\mathbf{x}^f|G_j) \mathbf{a}_j} \\ \nonumber
    & \approx \frac{p(\mathbf{x}^f|G_n) \mathbf{a}_n}{p(\mathbf{x}^f|G_n) \mathbf{a}_n + 0} = 1.
\end{align}
Accordingly, the semantic expectation at this point reduces to $\mathbf{e}(\mathbf{x}^f; \mathcal{G}) \approx \tilde{\mathbf{c}}_n$, with the learned opacity $\mathbf{a}_n$ nullified by the posterior normalization.

This issue is further exacerbated when considering the geometry prediction, where the opacity $\mathbf{a}_i$ is decomposed and depends solely on the Gaussian kernel. As such, even a low-confidence outlier can yield a high occupancy value for nearby points. Consequently, the voxel $\mathbf{x}^f$ is likely to be incorrectly activated as occupied by the semantic label of the outlier $G_n$, producing the characteristic ``floaters''.

\begin{figure}[t]
\centering
\includegraphics[width=0.85\columnwidth]{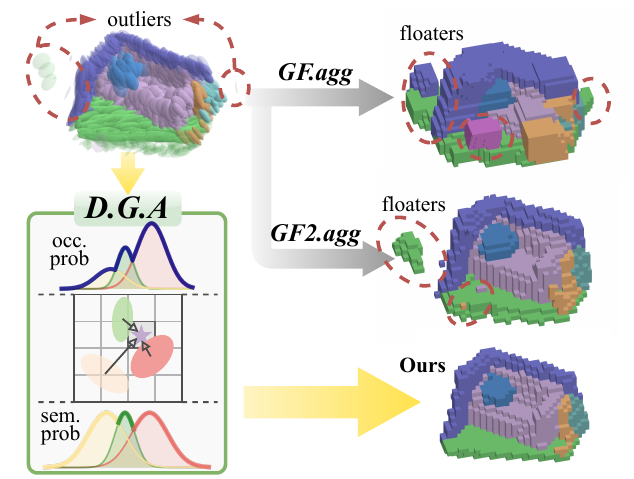} 
\caption{Illustration of the proposed DGA. While \textit{GF.agg}~\cite{huang2024gaussianformer} and \textit{GF2.agg}~\cite{huang2025gaussianformer2} wrongly produces the ``floaters'' from outliers, our DGA remains robust, as the low occupancy probability directly suppresses its erroneous semantic contribution. }
\label{fig4}
\vspace{-10pt}
\end{figure}

\subsubsection{Geometric Occupancy Prediction.}
Due to the exponential decay of the Gaussian kernel, only primitives in the local vicinity of $\mathbf{x}$ have a meaningful influence. Therefore, we only consider contributions from a neighborhood of relevant Gaussian primitives for efficiency, denoted as $\mathcal{N}(\mathbf{x})$. The occupancy is then modeled as a probabilistic OR operation over this local set. Crucially, each primitive's influence is modulated by its learned opacity $\mathbf{a}_i$, which we interpret as its existence confidence. This explicit use of opacity is a key difference from PGS: 
\begin{equation}
    \alpha'(x) = 1 - \prod_{i \in \mathcal{N}(\mathbf{x})} \left(1 - \alpha(\mathbf{x};G_i) \cdot \mathbf{a}_i\right).
\end{equation}
This natural gating mechanism suppresses the influence of low-confidence outliers on the final occupancy probability.
 
\subsubsection{Conditional Semantic Distribution.}
Concurrently, we predict the conditional semantic distribution $e(\mathbf{x})$ under the assumption that the position $\mathbf{x}$ is occupied. This is achieved by using GMM, where we leverage the normalized semantic weights of each Gaussian component. This design decouples the semantic prediction from the opacity parameter $\mathbf{a}_i$, forcing the model to rely solely on the geometric proximity and the learned softmax-normalized semantic properties $\tilde{\mathbf{c}}_i$ of each primitive. The posterior probability for each semantic class $k$ is then computed as: 
\vspace{-3pt}
\begin{equation} 
    e^k(\mathbf{x}) = \sum_{i \in \mathcal{N}(\mathbf{x})} p(G_i|\mathbf{x}) = \frac{\sum_{i \in \mathcal{N}(\mathbf{x})} p(\mathbf{x}|G_i) \cdot \tilde{\mathbf{c}}_i^k}{\sum_{j \in \mathcal{N}(\mathbf{x})} p(\mathbf{x}|G_j)}.
\end{equation}

\subsubsection{Probabilistic Fusion.}
Finally, the two decoupled pathways are fused to compute the final probability distribution $\hat{\mathbf{y}}_x$ for each 3D position $\mathbf{x}$. The probabilities for each valid semantic class $k$ and the empty class are defined as:  
\begin{align}
    \left\{
    \begin{array}{l} 
        \hat{\mathbf{y}}_x^k = \alpha'(\mathbf{x}) \cdot e^k(\mathbf{x}) \\
        \hat{\mathbf{y}}_x^{\mathrm{empty}} = 1 - \alpha'(\mathbf{x})
    \end{array}
    \right. .
\end{align}
This formulation serves as a principled and fully differentiable gating mechanism. A low occupancy probability $\alpha'(\mathbf{x})$, often resulting from an outlier primitive, directly suppresses any erroneous semantic prediction $e^k(\mathbf{x})$, thus elegantly eliminating ``floaters'' without complex heuristics. We demonstrate this effect in Figure~\ref{fig4}.

\subsection{Training Objective} 
Our model is trained via a two-stage strategy, where the first stage establishes a robust geometric prior before training the full network end-to-end. Throughout both stages, the pretrained model \textit{Depth-Anything-V2} is kept frozen. 

\subsubsection{Stage 1: Depth Branch Pre-training.}
In this stage, we exclusively train our depth branch to produce a high-quality geometry prior. This module is supervised by a composite depth loss $\mathcal{L}_{\mathrm{d}}$, similar with prior works~\cite{laina2016deeper, wang2025vggt}:
\begin{equation} \label{eq:loss_depth}
    \mathcal{L}_{\mathrm{d}} = \lambda_1 \mathcal{L}_{\mathrm{huber}}^{\mathrm{depth}} + \lambda_2 \mathcal{L}_{\mathrm{huber}}^{\mathrm{pts}} + \lambda_3 \mathcal{L}_{\mathrm{grad}}, 
\end{equation} 
where the terms are the depth Huber loss, point cloud Huber loss, and gradient matching Huber losses. 

\subsubsection{Stage 2: End-to-End SplatSSC Training.} 
In this stage, we train the entire SplatSSC network. To prevent the model from being overly constrained by the initial depth predictions, while maintaining a robust geometric prior, we remove $\mathcal{L}_{\mathrm{d}}$ and introduce our proposed \textit{Probability Scale Loss} \begin{small}$\mathcal{L}_{\mathrm{scal}}^{\mathrm{prob}}$\end{small} as a soft geometric supervision. The training objective is therefore optimized with a final composite loss $\mathcal{L}_{\mathrm{ssc}}$: 
\begin{equation} \label{eq:loss_total}
    \mathcal{L}_{\mathrm{ssc}} = \mathcal{L}_{\mathrm{sem}} + \lambda_4 \mathcal{L}_{\mathrm{scal}}^{\mathrm{prob}}, 
\end{equation} 
where $\mathcal{L}_{\mathrm{sem}} = \lambda_5 \mathcal{L}_{\mathrm{focal}} + \lambda_6 \mathcal{L}_{\mathrm{lovasz}}$ is the primary semantic segmentation loss adopted by EmbodiedOcc. Our loss \begin{small}$\mathcal{L}_{\mathrm{scal}}^{\mathrm{prob}}$\end{small} extends the geometry-aware scale loss \begin{small}$\mathcal{L}_{\mathrm{scal}}^{\mathrm{geo}}$\end{small} from MonoScene~\cite{cao2024monoscene}, adapting it to supervise the predicted occupancy probability across all $n$ encoder layers. To account for the progressive refinement across stages, we introduce a linear weighting schedule, which imposes weaker constraints on early-stage predictions and gradually enforces stronger consistency at deeper layers: 
\begin{equation} \label{eq:loss_prob}
    \mathcal{L}_{\mathrm{scal}}^{\mathrm{prob}} = \frac{1}{2} \sum_{i=1}^{n-1} \frac{i}{n} \cdot \mathcal{L}_{\mathrm{scal}}^{\mathrm{geo},i} + \mathcal{L}_{\mathrm{scal}}^{\mathrm{geo},n},
\end{equation}
where $i$ is the layer index. The loss weights are set as $\lambda_1=10$, $\lambda_2=20$, $\lambda_3=\lambda_4=0.5$, $\lambda_5=100$, and $\lambda_6=2$.

\begin{table*}[t]
\small
\centering
\setlength{\tabcolsep}{0.85mm}
\begin{tabular}{l|c|c|c|ccccccccccc|c}
\toprule 
\textbf{Dataset} & \textbf{Method} & \textbf{Input} & \textbf{IoU} & \rotatebox{90}{\textbf{ceiling}} & \rotatebox{90}{\textbf{floor}} & \rotatebox{90}{\textbf{wall}} & \rotatebox{90}{\textbf{window}} & \rotatebox{90}{\textbf{chair}} & \rotatebox{90}{\textbf{bed}} & \rotatebox{90}{\textbf{sofa}} & \rotatebox{90}{\textbf{table}} & \rotatebox{90}{\textbf{tvs}} & \rotatebox{90}{\textbf{furniture}} & \rotatebox{90}{\textbf{objects}} & \textbf{mIoU} \\
\midrule
\multirow{9}{*}{Occ-ScanNet} 
& TPVFormer & $\mathcal{I}_\mathrm{rgb}$ & 33.39 & 6.96 & 32.97 & 14.41 & 9.10 & 24.01 & 41.49 & 45.44 & 28.61 & 10.66 & 35.37 & 25.31 & 24.94 \\ 
& GaussianFormer & $\mathcal{I}_\mathrm{rgb}$ & 40.91 & 20.70 & 42.00 & 23.40 & 17.40 & 27.00 & 44.30 & 44.80 & 32.70 & 15.30 & 36.70 & 25.00 & 29.93 \\ 
& MonoScene & $\mathcal{I}_\mathrm{rgb}$ & 41.60 & 15.17 & 44.71 & 22.41 & 12.55 & 26.11 & 27.03 & 35.91 & 28.32 & 6.57 & 32.16 & 19.84 & 24.62 \\
& ISO & $\mathcal{I}_\mathrm{rgb}$ & 42.16 & 19.88 & 41.88 & 22.37 & 16.98 & 29.09 & 42.43 & 42.00 & 29.60 & 10.62 & 36.36 & 24.61 & 28.71 \\
& SurroundOcc & $\mathcal{I}_\mathrm{rgb}$ & 42.52 & 18.90 & 49.30 & 24.80 & 18.00 & 26.80 & 42.00 & 44.10 & 32.90 & 18.60 & 36.80 & 26.90 & 30.83 \\ 
& EmbodiedOcc & $\mathcal{I}_\mathrm{rgb}$ & 53.95 & 40.90 & 50.80 & 41.90 & 33.00 & 41.20 & 55.20 & 61.90 & 43.80 & 35.40 & 53.50 & 42.90 & 45.48 \\
& EmbodiedOcc++ & $\mathcal{I}_\mathrm{rgb}$ & 54.90 & 36.40 & 53.10 & 41.80 & 34.40 & 42.90 & \underline{57.30} & \underline{64.10} & 45.20 & 34.80 & 54.20 & 44.10 & 46.20 \\
& RoboOcc & $\mathcal{I}_\mathrm{rgb}$ & \underline{56.48} & \underline{45.36} & \underline{53.49} & \underline{44.35} & \underline{34.81} & \underline{43.38} & 56.93 & 63.35 & \underline{46.35} & \underline{36.12} & \underline{55.48} & \underline{44.78} & \underline{47.67} \\

& SplatSSC (Ours) & $\mathcal{I}_\mathrm{rgb}$ & \textbf{62.83} & \textbf{49.10} & \textbf{59.00} & \textbf{48.30} & \textbf{38.80} & \textbf{47.40} & \textbf{62.40} & \textbf{67.00} & \textbf{49.50} & \textbf{42.60} & \textbf{60.70} & \textbf{45.40} & \textbf{51.83} \\

\midrule
\multirow{5}{*}{Occ-ScanNet-mini} 
& MonoScene & $\mathcal{I}_\mathrm{rgb}$ & 41.90 & 17.00 & 46.20 & 23.90 & 12.70 & 27.00 & 29.10 & 34.80 & 29.10 & 9.70 & 34.50 & 20.40 & 25.90 \\
& ISO & $\mathcal{I}_\mathrm{rgb}$ & 42.90 & 21.10 & 42.70 & 24.60 & 15.10 & 30.80 & 41.00 & 43.30 & 32.20 & 12.10 & 35.90 & 25.10 & 29.40 \\
& EmbodiedOcc & $\mathcal{I}_\mathrm{rgb}$ & 55.13 & \underline{29.50} & 49.40 & 41.70 & 36.30 & 41.90 & 60.40 & 59.60 & 46.30 & \underline{34.50} & 58.00 & 43.50 & 45.57 \\
& EmbodiedOcc++ & $\mathcal{I}_\mathrm{rgb}$ & \underline{55.70} & 23.30 & \underline{51.00} & \underline{42.80} & \underline{39.30} & \underline{43.50} & \textbf{65.60} & \textbf{64.00} & \textbf{50.70} & \textbf{40.70} & \underline{60.30} & \textbf{48.90} & \underline{48.20} \\

& SplatSSC (Ours) & $\mathcal{I}_\mathrm{rgb}$ & \textbf{61.47} & \textbf{36.60} & \textbf{55.70} & \textbf{46.50} & \textbf{40.10} & \textbf{45.60} & \underline{64.50} & \underline{62.40} & \underline{48.60} & 30.60 & \textbf{61.20} & \underline{45.39} & \textbf{48.87} \\

\bottomrule
\end{tabular}%
\caption{Local Prediction Performance on the Occ-ScanNet dataset. The best results are highlighted in \textbf{bold}, while the second-best are \underline{underlined}.}\label{tab:occ_scannet_performance}
\end{table*}

\section{Experiments}

To evaluate the effectiveness of our SplatSSC, we conduct extensive experiments on the high-quality indoor datasets Occ-ScanNet and Occ-ScanNet-mini~\cite{yu2024monocular}. Details about datasets, implementation, and evaluation metrics are included in our supplementary material from our code.

\begin{table}[t]
\small
\centering 
\setlength{\tabcolsep}{1mm}
\begin{tabular}{cc|cc|c|cc}
\toprule  
Number & Scale Range & \makecell{Mem.$\downarrow$\\(MiB)} & \makecell{Time$\downarrow$\\(ms)} & Train & IoU & mIoU \\
\midrule 
19200 & [0.01, 0.08] & 3.122 & 135.18 & \checkmark & \textbf{62.77} & 47.69 \\ 
19200 & [0.01, 0.16] & 4.978 & 134.25 & \checkmark  & 60.64 & 43.31 \\ 
19200  & [0.01, 0.32] & 14.380 & 134.51 & OOM & / & / \\
4800  & [0.01, 0.08] & 3.158 & 123.27 & \checkmark & \underline{62.23} & 47.20 \\
4800  & [0.01, 0.16] & 3.108 & 122.63 & \checkmark & 61.53 & 46.74 \\
4800  & [0.01, 0.32] & 5.854 & 122.70 & \checkmark & 60.78 & 46.96 \\
1200  & [0.01, 0.08] & 3.104 & 116.20 & \checkmark & 60.18 & \underline{48.32} \\ 
1200  & [0.01, 0.16] & 3.112 & 115.56 & \checkmark & 61.47 & \textbf{48.87} \\
1200  & [0.01, 0.32] & 3.126 & 114.75 & \checkmark & 57.09 & 42.38 \\
\bottomrule 
\end{tabular}%
\caption{Ablation on Gaussian Parameters. Memory (Mem.) usage and time are measured on one 3090 GPU.}
\label{tab:ablation_gaussian_parameters}
\end{table}

\begin{table}[t]
\small
\centering
\setlength{\tabcolsep}{1mm}
\begin{tabular}{c|ccc|cc} 
\toprule
GMF & GF.agg& GF2.agg& DGA & IoU & mIoU \\ 
\midrule
- & \checkmark & - & - & 11.64 & 12.62 \\ 
- & - & \checkmark & - & 27.54 & 17.27 \\ 
- & - & - & \checkmark & 48.85 & 36.91 \\ 
\midrule
\checkmark & \checkmark & - & - & 16.63 & 10.45 \\
\checkmark & - & \checkmark & - & 57.70 & 45.13\\
\checkmark & - & - & \checkmark & \textbf{60.61} & \textbf{48.01}\\
\bottomrule 
\end{tabular}%
\caption{Ablation on the Components of SplatSSC.}
\label{tab:ablation_components}
\vspace{-10pt}
\end{table}

\subsection{Main Result} 
\begin{figure}[t]
\centering
\includegraphics[width=0.95\columnwidth]{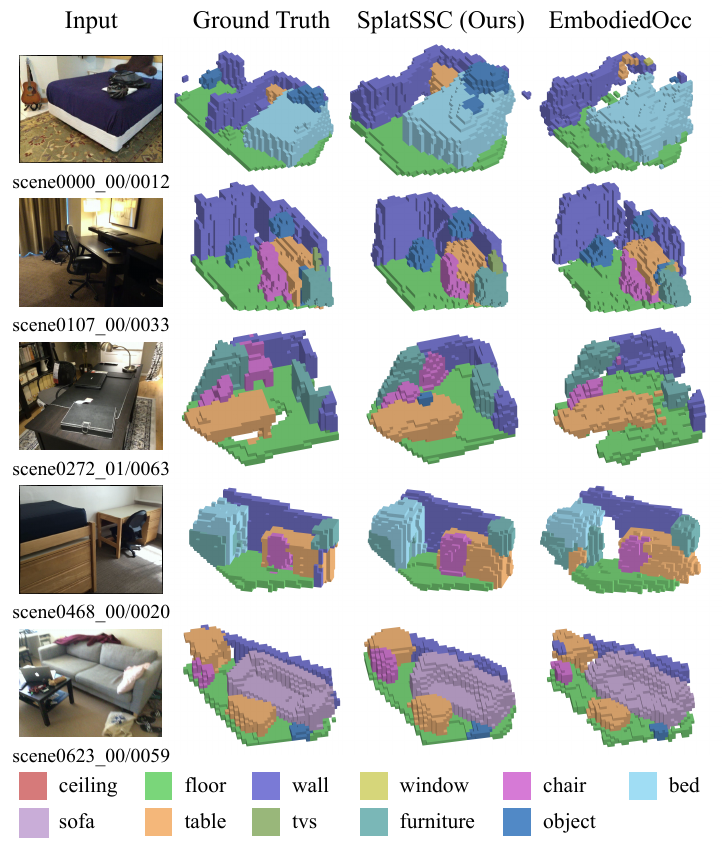}
\caption{Qualitative results on the Occ-ScanNet-mini dataset. Our method achieves superior performance in scene completion and target object recall compared to others.}
\label{fig:vis_occ}
\vspace{-10pt}
\end{figure}

The main results on the Occ-ScanNet and Occ-ScanNet-mini benchmarks are summarized in Table~\ref{tab:occ_scannet_performance}. Our SplatSSC achieves SOTA performance, demonstrating strong robustness and fine-grained scene understanding on both benchmarks. For Occ-ScanNet, SplatSSC achieves 62.83\% IoU and 51.83\% mIoU, surpassing the previous SOTA RoboOcc~\cite{zhang2025roboocc} by a substantial margin of 6.35\% and 4.16\%, respectively. The per-class analysis further highlights the consistent improvements brought by SplatSSC across diverse categories, from large structural elements (e.g., \textit{walls} and \textit{floor}) to fine-grained objects (e.g., \textit{sofas} and \textit{chairs}). These results underscore the strength of our synergistic design. The depth-guided initialization facilitates accurate geometric reconstruction, while our DGA ensures sharp semantic boundaries. As illustrated in the qualitative examples in Figure~\ref{fig:vis_occ}, SplatSSC yields superior 3D scene perception capabilities that surpass the previous paradigm.

\begin{table}[t]
\small
\centering
\setlength{\tabcolsep}{1mm}
\begin{tabular}{cccccc|cc}
\toprule 
$\mathcal{L}_\mathrm{focal}$ & $\mathcal{L}_\mathrm{lovasz}$ & $\mathcal{L}_\mathrm{scal}^\mathrm{prob}$ & $\mathcal{L}_\mathrm{scal}^\mathrm{geo}$ & $\mathcal{L}_\mathrm{scal}^\mathrm{sem}$ & $\mathcal{L}_\mathrm{d}$ & IoU & mIoU \\
\midrule  
\checkmark & \checkmark & - & \checkmark & \checkmark & - & 57.55 & 46.13 \\ 
\checkmark & \checkmark & \checkmark & - & - & \checkmark & \underline{60.34} & 46.67 \\
\checkmark & \checkmark & \checkmark & - & \checkmark & - & 59.19 & \textbf{48.28} \\ 
\checkmark & \checkmark & \checkmark & - & - & - & \textbf{60.61} & \underline{48.01} \\
\bottomrule
\end{tabular}%
\caption{Ablation on Training Objective.}
\label{tab:ablation_loss_functions}
\vspace{-10pt}
\end{table}

\begin{table}[t]
\small
\centering
\setlength{\tabcolsep}{1mm}
\begin{tabular}{c|cc|ccc}
\toprule 
GMF & DAv2 & FT-DAv2 & $\delta_1$ $\uparrow$ & RMSE $\downarrow$ & C-$l_1$ $\downarrow$ \\
\midrule 
- & \checkmark & - & 0.075 & 50.314 & 1.996 \\
\checkmark & \checkmark & - & 0.981 & 4.944 & 0.182 \\ 
\midrule 
- & - & \checkmark & 0.984 & 3.891 & 0.164 \\ 
\checkmark & - & \checkmark & \textbf{0.993} & \textbf{2.977} & \textbf{0.112} \\ 
\bottomrule 
\end{tabular}%
\caption{Ablation on Depth Branch.}
\label{tab:ablation_depth_branch}
\vspace{-10pt}
\end{table}

\begin{figure}[t]
\centering 
\includegraphics[width=0.95\columnwidth]{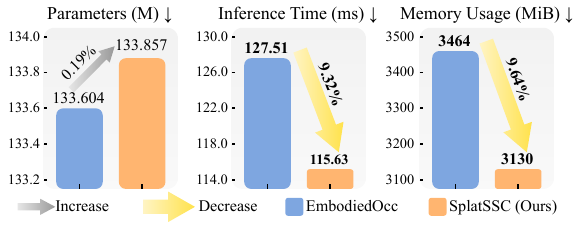}
\caption{Efficiency Analysis.}
\label{fig: efficiency}
\vspace{-10pt}
\end{figure}

\subsection{Ablation Studies} 
Ablation studies are conducted on the Occ-ScanNet-mini dataset to assess the impact of design choice in our model.
\subsubsection{Ablation on Gaussian Parameters.} 
We analyze the impact of primitive count and scale range in Table~\ref{tab:ablation_gaussian_parameters}, revealing a clear trade-off between performance and efficiency. Our setting achieves the highest semantic accuracy of $48.87\%$ mIoU with a remarkably compact configuration of just 1200 primitives. Increasing the count to 4800 and 19200 yields marginal gains in geometric completeness but incurs higher latency and lower mIoU. The choice of scale range is equally critical. Excessively large ranges degrade accuracy and trigger Out-of-Memory (OOM) failures under dense configurations, likely due to overlaps among oversized primitives. In contrast, a moderate range [0.01, 0.16] offers the best trade-off, effectively capturing both global layouts and fine-grained details with minimal redundancy.

\subsubsection{Ablation on Network Components.}
We evaluate the impact of our key components, GMF and DGA, in Table~\ref{tab:ablation_components}.  
The analysis first highlights the necessity of a tailored aggregation mechanism. The standard GF.agg~\cite{huang2024gaussianformer} nearly fails in our sparse setting, yielding a prohibitively low 10.45\% mIoU. While the more advanced GF2.agg~\cite{huang2025gaussianformer2} performs significantly better, our DGA still surpasses it by over $2.8\%$ in both IoU and mIoU. This confirms that ``floaters'' are the key bottleneck in sparse splatting, and DGA is crucial for efficient and robust aggregation.  
The proposed GMF is equally important, as replacing it with a naive depth-aware baseline~\cite{wu2024embodiedocc} built on Depth-Anything-V2 causes a substantial drop by more than $11\%$ in both geometries and semantics, even when paired with our DGA. The degradation becomes more severe with other aggregators, leading to a near-collapse in performance. This demonstrates the necessity of structured geometric priors for generating informative primitives.

\subsubsection{Ablation on Training Objective.}
Our validation on the training objective design is shown in Table~\ref{tab:ablation_loss_functions}. The results first confirm that the popular combination of geometry and semantic scale losses ($\mathcal{L}_{\mathrm{scal}}^{\mathrm{geo}}$, $\mathcal{L}_{\mathrm{scal}}^{\mathrm{sem}}$) is suboptimal for our framework, yielding the lowest 46.13\% mIoU. The explicit depth loss $\mathcal{L}_d$ is also detrimental, as its inclusion consistently degrades both geometric and semantic scores. Furthermore, while adding semantic scale loss provides a marginal mIoU boost to a peak of 48.28\%, it incurs over 1.42\% IoU drop. These findings lead to our final design: a simple yet effective objective incorporating our proposed \begin{small}$\mathcal{L}_{\mathrm{scal}}^{\mathrm{prob}}$\end{small} alongside standard Focal and Lovász losses ($\mathcal{L}_\mathrm{focal}$, $\mathcal{L}_\mathrm{lovasz}$), which achieves the best geometric performance of 60.61\% IoU while maintaining competitive semantic accuracy.
 
\subsubsection{Ablation on Depth Branch.}
The contribution of our GMF module is validated in Table~\ref{tab:ablation_depth_branch}. The results highlight a dramatic impact of GMF on refining the geometric prior. When applied to a frozen Depth-Anything-V2 (DAv2) backbone, GMF boosts the $\delta_1$ score by a remarkable 0.906. Furthermore, it demonstrates its capability to enhance a fine-tuned Depth-Anything-V2 (FT-DAv2)~\cite{wu2024embodiedocc}, pushing the $\delta_1$ score to a new best of 0.993. This confirms that our GMF is a powerful and versatile feature refiner, essential for generating high-quality geometric representations.

\subsubsection{Efficiency Analysis.}
Beyond accuracy, we evaluate the computational efficiency of SplatSSC against EmbodiedOcc, with results detailed in Figure~\ref{fig: efficiency}. Our method demonstrates superior efficiency despite a negligible $0.19\%$ increase in parameter count. Specifically, SplatSSC achieves a $9.32\%$ reduction in inference latency and a $9.64\%$ decrease in memory usage. This advantage is primarily attributed to our sparse design, which operates on significantly fewer primitives than prior works.

\section{Conclusion}
In this paper, we introduced SplatSSC, a novel framework for monocular 3D semantic scene completion. Our method addresses the critical limitations of prior object-centric approaches through two core technical contributions: 1) a depth-guided initialization strategy, powered by our group-wise multi-scale fusion module, which generates a compact and high-quality set of initial Gaussian primitives; and 2) a decoupled Gaussian aggregator that robustly resolves aggregation artifacts such as ``floaters'' from outlier primitives. Extensive experiments demonstrate that SplatSSC establishes a new SOTA on the Occ-ScanNet benchmark, achieving superior accuracy while simultaneously reducing latency and memory consumption.

Despite its outstanding performance, we acknowledge several limitations that offer avenues for future work as discussed in the supplementary material from our code.

\section{Acknowledgments}
This work was supported by National Research Foundation of Singapore Medium-sized Centre for Advanced Robotics Technology Innovation and Ministry of Education, Singapore, under AcRF TIER 1 Grant RG64/23.

\input{appendix}

\bibliography{aaai2026}

\end{document}

%% file: appendix.tex
\section{Appendix Overview}
This technical appendix consists of the following sections.
\begin{itemize}
    \item We detail the experimental setup for SplatSSC.
    \item We provide a detailed derivation of the semantic probability formulation for our proposed Decoupled Gaussian Aggregator (DGA).
    \item We provide further visualization of qualitative results on the Occ-ScanNet-mini and Occ-ScanNet validation sets.
    \item We provide further proof and analysis to support the necessity of our two-stage training.
    \item We conclude with a discussion of the current limitations and potential applications of SplatSSC.
    \item We include a statement regarding our code availability and its license.
\end{itemize}

\section{Experimental Setup}
\subsection{Dataset} 
Occ-ScanNet~\cite{yu2024monocular} comprises 45,755 training frames and 19,764 validation frames, annotated with 12 semantic classes, with one representing free space and eleven corresponding to specific categories, including ceiling, floor, wall, window, chair, bed, sofa, table, television, furniture, and generic objects. The ground truth is provided as a voxel grid covering a 4.8m$\times$4.8m$\times$2.88m region in front of the camera, discretized into a resolution of 60$\times$60$\times$36. This dataset serves as the benchmark for training and evaluating local occupancy prediction. A smaller variant, Occ-ScanNet-mini, is also available, containing 5,504 training and 2,376 validation frames.

\subsection{Evaluation metrics}
Following common practice~\cite{cao2024monoscene, hu2024metric3d}, we evaluate the final semantic scene completion performance using Intersection-over-Union (IoU) and mean IoU (mIoU). These metrics are computed exclusively within the current camera's view frustum. To assess the quality of the geometric prediction in our depth branch, we employ three additional metrics: Chamfer $l_1$ distance (C-$l_1$), Root Mean Squared Error (RMSE), and accuracy under threshold ($\delta_1$)~\cite{hu2024metric3d}. For this geometric evaluation, the ground truth point cloud is generated by down-sampling the ground truth depth map using the indices from our GMF module, then projecting the valid depth points into the camera's coordinate space.
 
\subsection{Implementation Details}
In our framework, the image encoder employs a pretrained EfficientNet-B7~\cite{tan2019efficientnet} as backbone, while the depth branch utilizes a frozen fine-tuned \textit{Depth-Anything-V2}~\cite{wu2024embodiedocc} model. For both training stages, we use the AdamW optimizer~\cite{loshchilov2017decoupled} with a weight decay of 0.01. We apply a learning rate multiplier of 0.1 to the backbone. All input images are processed at a resolution of $480 \times 640$.

\subsubsection{Stage 1: Depth Branch Pretraining.}
In the first stage, we exclusively pretrain our depth branch to establish a robust geometric prior. The down-sampled grid for our GMF has a shape of $30 \times 40$. We employ a cosine learning rate schedule with a 1000-iteration warmup, setting the peak learning rate to $6 \times 10^{-4}$. The model is trained for 10 epochs on the Occ-ScanNet dataset using 2 NVIDIA 3090 GPUs with a per-GPU batch size of 2 (total batch size of 4).

\subsubsection{Stage 2: End-to-End SplatSSC Training.}
In the second stage, we train the full SplatSSC model, initializing the depth branch with weights from stage one. The $30 \times 40$ down-sampled grid generates an initial set of 1200 Gaussian primitives, with their scales initialized in the range [0.01m, 0.16m]. We train the model on 4 NVIDIA 4090 GPUs with a per-GPU batch size of 2, resulting in a total batch size (bs) of 8. The learning rate follows a cosine schedule with a 1000-iteration warmup, and the peak learning rate is determined by a linear scaling rule: $2 \times 10^{-4} \cdot (\text{bs}/2)$. The model is trained for 10 epochs on the full Occ-ScanNet dataset and for 20 epochs on the Occ-ScanNet-mini subset.

\subsubsection{Further experiments settings.} 
The experimental settings for the ablation studies and efficiency analysis are summarized in Table~\ref{tab: ablation_config}. 

\begin{table*}[t]
\small
\centering
\setlength{\tabcolsep}{1mm}
\begin{tabular}{l|cccc|c}
\toprule
\multirow{2}{*}{Config} & \multicolumn{4}{c|}{Ablation Studies} & \multirow{2}{*}{Efficiency Analysis} \\
& Gaussian Parameters & Components of SplatSSC & Training Objective & Depth Branch & \\
\midrule 
Training Dataset & Occ-ScanNet-mini & Occ-ScanNet-mini & Occ-ScanNet-mini & Occ-ScanNet & Occ-ScanNet \\
Inference Dataset & Occ-ScanNet-mini & Occ-ScanNet-mini & Occ-ScanNet-mini & Occ-ScanNet-mini & Occ-ScanNet \\
Training Device & 4 RTX 3090 & 2 RTX 3090 & 2 RTX 3090 & 2 RTX 3090 & 4 RTX 4090 \\ 
Inference Device & 1 RTX 3090 & 1 RTX 3090 & 1 RTX 3090 & 1 RTX 3090 & 1 RTX 3090  \\ 
Maximum Learning Rate & $8\times 10^{-4}$ & $6\times 10^{-4}$ & $6\times 10^{-4}$ & $6\times 10^{-4}$ & $8\times 10^{-4}$ \\
Weight Decay & 0.01 & 0.01 & 0.01 & 0.01 & 0.01 \\ 
Total Batch Size & 8 & 6 & 6 & 6 & 8 \\ 
\bottomrule 
\end{tabular} 
\caption{Experiment settings for different ablation studies and efficient analysis.} 
\label{tab: ablation_config} 
\end{table*} 
   
\section{Derivation of Decoupled Gaussian Aggregator}
\label{sec:appendix_aggregator}
This section presents a complete derivation of the proposed Decoupled Gaussian Aggregator (DGA), clarifying the probabilistic reasoning behind the semantic term that models the probability of a point $\mathbf{x}$ belonging to class $k$, given that it is occupied.

For clarity, we restate the definition of Gaussian primitives, $\mathcal{G}=\{G_i\}^N_{i=1}$, with each Gaussian parameterized by a mean $\boldsymbol{\mu}_i \in \mathbb{R}^3$, a scale vector $\mathbf{s}_i \in \mathbb{R}^3$, a rotation quaternion $\mathbf{q}_i \in \mathbb{R}^4$, a learned opacity $\mathbf{a}_i \in [0, 1]$, and a softmax-normalized semantic vector $\tilde{\mathbf{c}}_i \in \mathbb{R}^{C}$.

Our DGA is designed to explicitly separate the prediction of geometry and semantics. While define the final prediction as \begin{small}$\hat{\mathbf{y}}^k(\mathbf{x}) = \alpha'(\mathbf{x}) \cdot e^k(\mathbf{x})$\end{small} for valid classes, we incorporate opacities $\mathbf{a}_i$ into the occupancy probability $\alpha'(\mathbf{x})$ and formulate a conditional semantic distribution $e^k(\mathbf{x})$. 

We model the semantic distribution as a Gaussian mixture model, where each primitive $G_i$ in a local neighborhood $\mathcal{N}(\mathbf{x})$ is a component. The likelihood of $G_i$ contributing to class $k$ is determined by its semantic affinity $\tilde{\mathbf{c}}_i^k$. Following this, we can formulate the semantic probability for class $k$ at point $\mathbf{x}$ using Bayes' theorem:
\begin{equation}
    e^k(\mathbf{x}) = \frac{\sum_{i \in \mathcal{N}(\mathbf{x})} p(\mathbf{x} | G_i) \tilde{\mathbf{c}}_i^k}{\sum_{j \in \mathcal{N}(\mathbf{x})} \sum_{l=1}^{C} p(\mathbf{x} | G_j) \tilde{\mathbf{c}}_j^l}.
\end{equation}
This initial expression can be further simplified. By factoring out the likelihood term in the denominator part, we have:
\begin{equation}
    \sum_{j \in \mathcal{N}(\mathbf{x})} \sum_{l=1}^{C} p(\mathbf{x} | G_j) \tilde{\mathbf{c}}_j^l = \sum_{j \in \mathcal{N}(\mathbf{x})} p(\mathbf{x} | G_j) \left( \sum_{l=1}^{C} \tilde{\mathbf{c}}_j^l \right).
\end{equation}
As the semantic vector $\tilde{\mathbf{c}}_j$ is softmax-normalized, the sum of its components over all classes is unity, i.e., \begin{small}$\sum_{l=1}^{C} \tilde{\mathbf{c}}_j^l = 1$\end{small}. This crucial property simplifies the normalization term to the sum of only the geometric likelihoods.

Substituting this back, we arrive at the final expression for our conditional semantic distribution:
\begin{equation}
    e^k(\mathbf{x}) = \frac{\sum_{i \in \mathcal{N}(\mathbf{x})} p(\mathbf{x} | G_i) \tilde{\mathbf{c}}_i^k}{\sum_{j \in \mathcal{N}(\mathbf{x})} p(\mathbf{x} | G_j)}.
\end{equation}

\begin{figure}[t]  
\centering
\includegraphics[width=0.95\columnwidth]{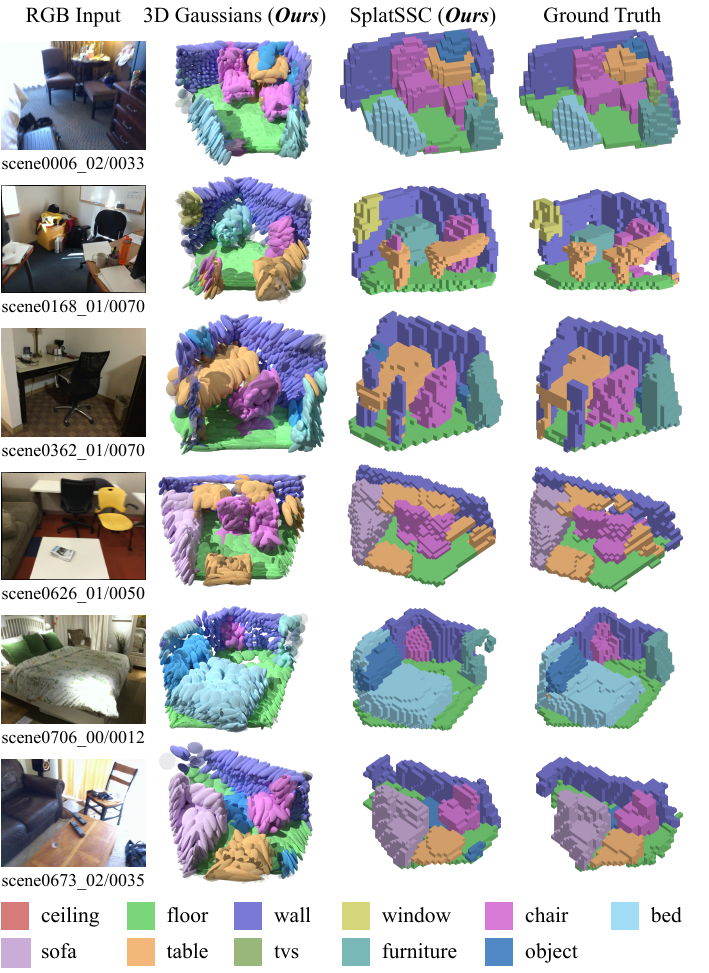} %
\caption{Further visualization on Occ-ScanNet-mini.}
\label{fig:vis_mini}
\end{figure} 

\begin{figure}[t]  
\centering
\includegraphics[width=0.95\columnwidth]{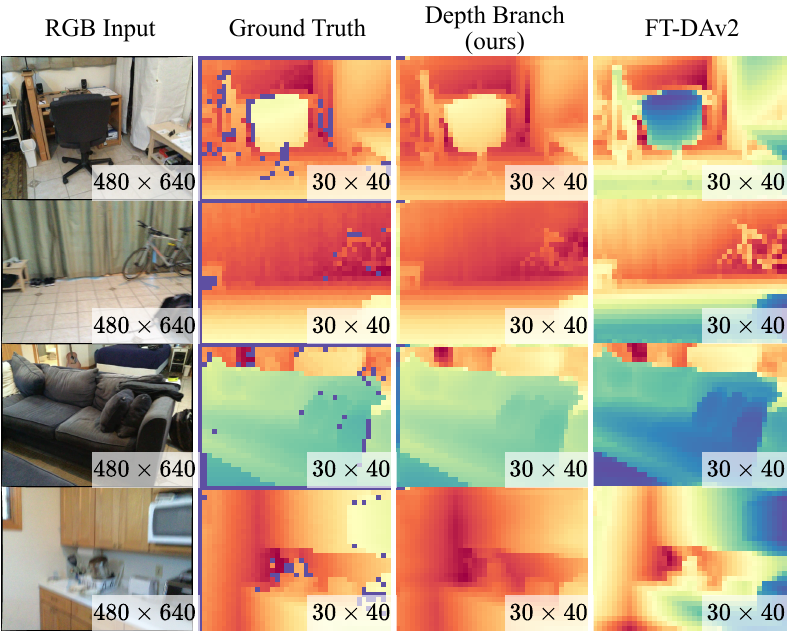} %
\caption{Qualitative visualization on the depth prediction.}
\label{fig:depth_compare}
\end{figure}

\begin{figure}[t]  
\centering
\includegraphics[width=0.95\columnwidth]{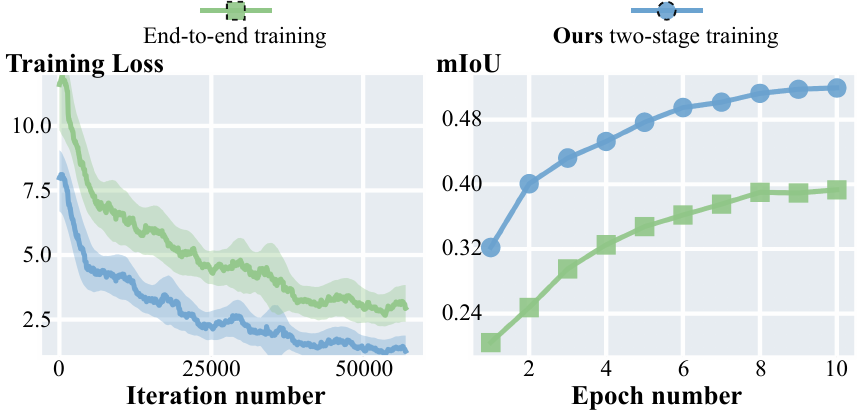} %
\caption{Comparison of different training strategies.} 
\label{fig:training_strategy}
\end{figure}
\textbf{}

\section{Additional Visualization Results}
\subsection{Further Visualization on SSC}
\begin{figure*}[t]
\centering
\includegraphics[width=0.95\textwidth]{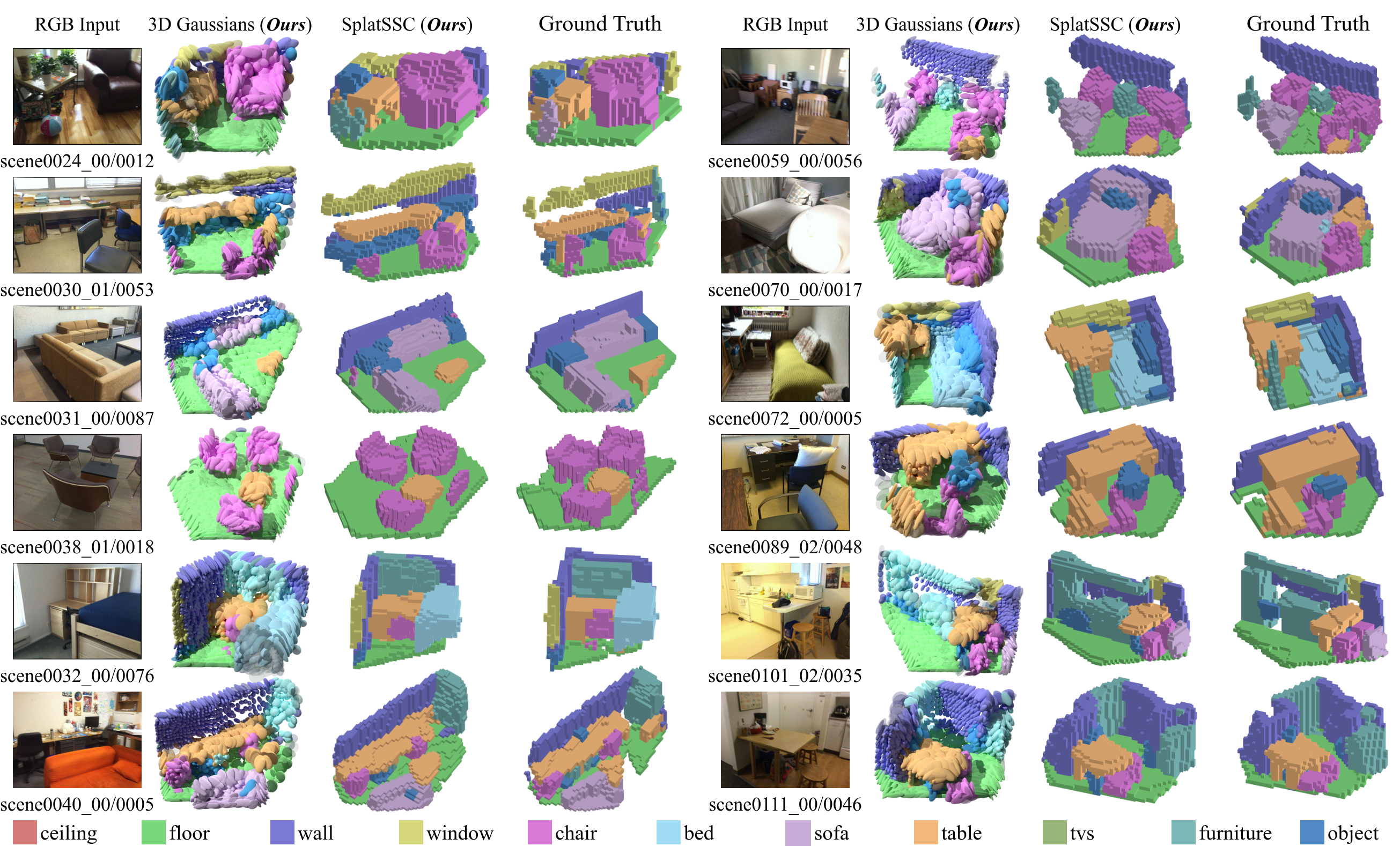} %
\caption{Further visualization of predicted 3D Gaussians and voxels on the Occ-ScanNet dataset.}
\label{fig:vis_base}
\end{figure*}

In Figure~\ref{fig:vis_base}, we present additional qualitative results on the Occ-ScanNet validation set. For each scene, we visualize both the intermediate 3D semantic Gaussians and the final dense occupancy prediction. The Gaussian views reveal that SplatSSC forms a sparse, object-centric representation that already captures well-aligned shapes and semantics for major structures (e.g., furniture and walls), while the rendered voxel grids show that this compact representation can be faithfully translated into detailed 3D completions.

Furthermore, Figure~\ref{fig:vis_mini} provides more visualizations on the Occ-ScanNet-mini validation set using the same layout. Again, we show the per-frame 3D semantic Gaussians together with the final occupancy predictions. These examples highlight that our Gaussian scene representation remains stable and expressive in smaller and more cluttered environments, and that it consistently supports accurate reconstruction of fine-scale object geometry.

\subsection{Further Visualization on Depth Branch} 
Figure~\ref{fig:depth_compare} provides additional qualitative comparisons of our depth branch. We visualize $30 \times 40$ depth maps from our method and FT-DAv2 against downsampled ground truth, which serve as spatial priors for Gaussian primitive initialization. Compared to FT-DAv2, our branch produces depth maps with higher fidelity and sharper foreground-background separation, aligning more closely with the ground truth.

These results indicate that our proposed depth branch can effectively refine the depth prior into a more structured and compact geometric representation, providing a stronger initialization for the subsequent Gaussian lifting stage.

\section{Discussion} 

\subsection{Analysis of Two-Stage Training Strategy}
Figure~\ref{fig:training_strategy} compares our two-stage schedule with an end-to-end training variant. In both cases, we use the SSC loss:
\begin{align}
    \mathcal{L}_{\mathrm{ssc}} &= \mathcal{L}_{\mathrm{sem}} + \lambda_1 \mathcal{L}_{\mathrm{scal}}^{\mathrm{prob}}, \\
    \mathcal{L}_{\mathrm{sem}} &= \lambda_2 \mathcal{L}_{\mathrm{focal}} + \lambda_3 \mathcal{L}_{\mathrm{lovasz}}.
\end{align} 
Our two-stage model in the second stage is optimized only with $\mathcal{L}_{\mathrm{ssc}}$, while the end-to-end baseline additionally uses the depth loss: 
\begin{align}
\mathcal{L}_{\mathrm{d}} = \lambda_4 \mathcal{L}_{\mathrm{huber}}^{\mathrm{depth}} 
+ \lambda_5 \mathcal{L}_{\mathrm{huber}}^{\mathrm{pts}} 
+ \lambda_6 \mathcal{L}_{\mathrm{grad}}.
\end{align}
For a fair comparison, Figure~\ref{fig:training_strategy} reports only $\mathcal{L}_{\mathrm{ssc}}$ together with the validation mIoU. In our experiments, we set $\lambda_1=\lambda_6=0.5$, $\lambda_2=100$, $\lambda_3=2$, $\lambda_4=20$, $\lambda_5=10$.

We observe that the end-to-end variant converges to a larger $\mathcal{L}_{\mathrm{ssc}}$ and a lower mIoU, whereas our two-stage schedule achieves a smaller loss and a higher mIoU. We attribute this gap to the absence of a precise depth prior in the end-to-end setting: jointly training the depth branch from scratch couples noisy depth supervision with SSC optimization and makes the Gaussian lifting less stable. In contrast, the first stage of our two-stage training learns a reliable depth prior under metric supervision, and the second stage then refines the Gaussian representation using $\mathcal{L}_{\mathrm{ssc}}$ only, which better aligns the pretrained depth cues with the SSC objective.

\subsection{Limitations} 
Despite the strong performance, our SplatSSC framework has certain design constraints that highlight key areas for further improvement.
 
\subsubsection{Hyperparameter Sensitivity Analysis.} 
This experiment validates our finding that \textit{\textbf{SplatSSC's performance is subject to a distinct threshold}} regarding its training hyperparameters, with results shown in Table~\ref{tab:ablation_hyperparams}. This effect is visible when comparing performance at different batch sizes. At a total batch size of 2, our model's performance is substantially limited to 36.09\% mIoU. However, upon increasing the batch size to 4, the mIoU jumps dramatically to 45.32\%, reaching a competitive level. This demonstrates that a batch size of at least 4 is necessary for effective optimization. Beyond this threshold, performance continues to scale robustly, with the best results achieved at a batch size of 8. In contrast, the baseline EmbodiedOcc~\cite{wu2024embodiedocc} exhibits only modest and linear gains. It is also worth noting that EmbodiedOcc was designed for a per-GPU batch size of one, making extensive scaling less applicable. This highlights that the observed threshold effect is a unique characteristic of our model's interactive primitive optimization. 
 
\subsubsection{Local-View Architectural Constraint.} 
The current SplatSSC framework is designed to operate on a per-frame basis, excelling at generating a high-quality scene representation from a single view. However, this design presents a scalability challenge when extending to global scene perception. A naive extension of simply accumulating primitives from consecutive frames would lead to an unbounded growth in their total number, causing a rapid escalation in both memory and computational costs. This limitation reveals a critical need for a scalable online primitive management strategy that leverages both pruning and fusion techniques to prevent unbounded growth in memory and computation. We leave this as a promising direction for future work and will validate it on global scene benchmarks~\cite{wu2024embodiedocc, wang2024embodiedscan}.

\begin{table}[t]
\small
\centering
\setlength{\tabcolsep}{0.85mm}
\begin{tabular}{c|cc|cc}
\toprule
Method & bs & lr & IoU & mIoU \\
\midrule
\multirow{2}{*}{EmbodiedOcc} 
& 2 & 2$\times 10^{-4}$ & 52.59 & 42.61 \\
& 4 & 2$\times 10^{-4}$ & 55.13\,(+2.54\%) & 45.57\,(+2.96\%) \\
\midrule
\multirow{4}{*}{SplatSSC} 
& 2 & 2$\times 10^{-4}$ & 54.68 & 36.09 \\
& 4 & 4$\times 10^{-4}$ & 59.53\,(+4.85\%) & 45.32\,(+9.23\%) \\
& 6 & 6$\times 10^{-4}$ & 61.47\,(+6.79\%) & 48.87\,(+12.78\%) \\
& 8 & 8$\times 10^{-4}$ & 62.83\,(+8.15\%) & 51.83\,(+15.74\%) \\
\bottomrule
\end{tabular}%
\caption{Hyperparameter Sensitivity Analysis. We evaluate the performance on different total batch sizes (bs) and maximum learning rates (lr).}\label{tab:ablation_hyperparams}
\end{table}

\subsection{Future Outlook and Broader Applications}
While SplatSSC establishes a new state-of-the-art, its underlying principles open up several exciting avenues for future research. We discuss two key directions below.

\subsubsection{Scaling to Unbounded and Large-Scale Environments.}
A primary direction is adapting SplatSSC for large-scale outdoor environments, particularly for applications like autonomous driving. Unlike methods that rely on dense grids~\cite{wei2023surroundocc, zhang2023occformer} or random initialization across a predefined volume~\cite{huang2025gaussianformer2}, our depth-guided approach naturally focuses computation on observed surfaces. This inherent efficiency makes it exceptionally well-suited for sparse and large-scale settings. To fully realize this potential, the fixed volumetric grid could be replaced with more flexible spatial data structures, such as hash-encoded grids~\cite{deng2025vpgs}, to support unbounded scenes. This extension would also need to address the challenges unique to this domain, such as managing a dynamically growing set of primitives and handling the presence of dynamic objects.

\subsubsection{Application in Embodied AI and Robotics.}
Moving beyond passive perception, a critical frontier in 3D vision is to build representations that support active interaction, a central theme in embodied and spatial intelligence~\cite{wang2024embodiedscan, halacheva2025gaussianvlm}. 
Applying SplatSSC in embodied AI requires moving from single-frame perception to building a persistent and interactive world model. This demands a higher level of detail than is currently captured; for instance, an agent needs not just a semantic label for a ``door'', but also precise geometric information about its handle for manipulation. This may necessitate using a larger number of Gaussians or a finer-grained semantic taxonomy. Furthermore, it requires a robust online framework where the agent can continuously fuse new observations~\cite{deng2025mne}, prune outdated information, and refine its Gaussian-based world map in real-time. 

\section{Code Availability and Licensing}
The source code and trained models associated with this paper will be released on GitHub upon acceptance, and the specific URL will be provided. 

All our source code is licensed under the Creative Commons Attribution-NonCommercial-ShareAlike 4.0 International (CC BY-NC-SA 4.0) license. This permits any non-commercial use, distribution, and reproduction in any medium, provided the original work is properly cited and any derivative works are shared under the same license.

%% file: aaai2026.bib
@inproceedings{tong2023scene,
  author       = {Wenwen Tong and
                  Chonghao Sima and
                  Tai Wang and
                  Li Chen and
                  Silei Wu and
                  Hanming Deng and
                  Yi Gu and
                  Lewei Lu and
                  Ping Luo and
                  Dahua Lin and
                  Hongyang Li
                 },
  title        = {Scene as Occupancy},
  booktitle    = {Proceedings of the {IEEE/CVF} International Conference on Computer Vision},
  pages        = {8372--8381},
  publisher    = {{ICCV}},
  year         = {2023}
}

@inproceedings{huang2023tri,
  author       = {Yuanhui Huang and
                  Wenzhao Zheng and
                  Yunpeng Zhang and
                  Jie Zhou and
                  Jiwen Lu
                 },
  title        = {Tri-Perspective View for Vision-Based 3D Semantic Occupancy Prediction},
  booktitle    = {Proceedings of the {IEEE/CVF} Conference on Computer Vision and Pattern Recognition},
  pages        = {9223--9232},
  publisher    = {{CVPR}},
  year         = {2023}
}

@inproceedings{wei2023surroundocc,
  author       = {Yi Wei and
                  Linqing Zhao and
                  Wenzhao Zheng and
                  Zheng Zhu and
                  Jie Zhou and
                  Jiwen Lu
                 },
  title        = {SurroundOcc: Multi-Camera 3D Occupancy Prediction for Autonomous Driving},
  booktitle    = {Proceedings of the {IEEE/CVF} International Conference on Computer Vision},
  pages        = {21672--21683},
  publisher    = {{ICCV}},
  year         = {2023}
}

@inproceedings{tian2023occ3d,
  author       = {Xiaoyu Tian and
                  Tao Jiang and
                  Longfei Yun and
                  Yucheng Mao and
                  Huitong Yang and
                  Yue Wang and
                  Yilun Wang and
                  Hang Zhao
                 },
  title        = {Occ3D: {A} Large-Scale 3D Occupancy Prediction Benchmark for Autonomous Driving},
  booktitle    = {Advances in Neural Information Processing Systems},
  volume       = {36},
  pages        = {64318--64330},
  publisher    = {{NeurIPS}},
  year         = {2023}
}

@inproceedings{wang2024panoocc,
  author       = {Yuqi Wang and
                  Yuntao Chen and
                  Xingyu Liao and
                  Lue Fan and
                  Zhaoxiang Zhang
                 },
  title        = {PanoOcc: Unified Occupancy Representation for Camera-based 3D Panoptic Segmentation},
  booktitle    = {Proceedings of the {IEEE/CVF} Conference on Computer Vision and Pattern Recognition},
  pages        = {17158--17168},
  publisher    = {{CVPR}},
  year         = {2024}
}

@inproceedings{zhang2023occformer, 
  author={Zhang, Yunpeng and Zhu, Zheng and Du, Dalong},
  year={2023}, 
  title={Occformer: Dual-path transformer for vision-based 3d semantic occupancy prediction},
  booktitle={Proceedings of the {IEEE/CVF} International Conference on Computer Vision},
  pages={9433--9443}, 
  publisher={{ICCV}}, 
}

@misc{yu2023flashocc,
  title={Flashocc: Fast and memory-efficient occupancy prediction via channel-to-height plugin},
  author={Yu, Zichen and Shu, Changyong and Deng, Jiajun and Lu, Kangjie and Liu, Zongdai and Yu, Jiangyong and Yang, Dawei and Li, Hui and Chen, Yan}, 
  eprint={2311.12058},
  archivePrefix="arxiv", 
  year={2023}
}

@inproceedings{hou2024fastocc,
  author       = {Jiawei Hou and
                  Xiaoyan Li and
                  Wenhao Guan and
                  Gang Zhang and
                  Di Feng and
                  Yuheng Du and
                  Xiangyang Xue and
                  Jian Pu
                 },
  title        = {FastOcc: Accelerating 3D Occupancy Prediction by Fusing the 2D Bird's-Eye View and Perspective View},
  booktitle    = {{IEEE} International Conference on Robotics and Automation}, 
  pages        = {16425--16431}, 
  publisher    = {{ICRA}},
  year         = {2024}
}

@inproceedings{shi2024occupancy,
  author={Shi, Yiang and Cheng, Tianheng and Zhang, Qian and Liu, Wenyu and Wang, Xinggang},
  title={Occupancy as set of points},
  booktitle={Proceedings of the European Conference on Computer Vision},
  volume={15119},
  pages={72--87},
  publisher={{ECCV}},
  year={2024}
}

@inproceedings{wang2024opus,
  title={Opus: occupancy prediction using a sparse set},
  author={Wang, Jiabao and Liu, Zhaojiang and Meng, Qiang and Yan, Liujiang and Wang, Ke and Yang, Jie and Liu, Wei and Hou, Qibin and Cheng, Ming-Ming},
  booktitle={Advances in Neural Information Processing Systems},
  volume={37},
  pages={119861--119885},
  publisher={{NeurIPS}}, 
  year={2024}
}

@inproceedings{huang2025gaussianformer2,
  author       = {Yuanhui Huang and
                  Amonnut Thammatadatrakoon and
                  Wenzhao Zheng and
                  Yunpeng Zhang and
                  Dalong Du and
                  Jiwen Lu
                 },
  title        = {GaussianFormer-2: Probabilistic Gaussian Superposition for Efficient 3D Occupancy Prediction},
  booktitle    = {Proceedings of the {IEEE/CVF} Conference on Computer Vision and Pattern Recognition},
  pages        = {27477--27486},
  publisher    = {CVPR},
  year         = {2025}
}

@inproceedings{tang2024sparseocc,
  author       = {Pin Tang and
                  Zhongdao Wang and
                  Guoqing Wang and
                  Jilai Zheng and
                  Xiangxuan Ren and
                  Bailan Feng and
                  Chao Ma
                 },
  title        = {SparseOcc: Rethinking Sparse Latent Representation for Vision-Based Semantic Occupancy Prediction},
  booktitle    = {Proceedings of the {IEEE/CVF} Conference on Computer Vision and Pattern Recognition},
  pages        = {15035--15044},
  publisher    = {{CVPR}},
  year         = {2024}
}

@inproceedings{huang2024gaussianformer,
  author       = {Yuanhui Huang and
                  Wenzhao Zheng and
                  Yunpeng Zhang and
                  Jie Zhou and
                  Jiwen Lu
                 }, 
  title        = {GaussianFormer: Scene as Gaussians for Vision-Based 3D Semantic Occupancy Prediction},
  booktitle    = {Proceedings of the European Conference on Computer Vision},
  volume       = {15085},
  pages        = {376--393},
  publisher    = {{ECCV}},
  year         = {2024}
}

@misc{zhao2025gaussianformer3d,
  title={GaussianFormer3D: Multi-Modal Gaussian-based Semantic Occupancy Prediction with 3D Deformable Attention},
  author={Zhao, Lingjun and Wei, Sizhe and Hays, James and Gan, Lu},
  eprint={2505.10685},
  archivePrefix="arxiv",
  year={2025}
}

@article{kerbl20233d,
  author       = {Bernhard Kerbl and
                  Georgios Kopanas and
                  Thomas Leimk{\"{u}}hler and
                  George Drettakis
                 },
  title        = {3D Gaussian Splatting for Real-Time Radiance Field Rendering},
  journal      = {{ACM} Trans. Graph.},
  volume       = {42},
  number       = {4},
  pages        = {139:1--139:14},
  year         = {2023}
}

@misc{deng2025vpgs,
  title={VPGS-SLAM: Voxel-based Progressive 3D Gaussian SLAM in Large-Scale Scenes},
  author={Deng, Tianchen and Wu, Wenhua and He, Junjie and Pan, Yue and Jiang, Xirui and Yuan, Shenghai and Wang, Danwei and Wang, Hesheng and Chen, Weidong},
  eprint={2505.18992},
  archivePrefix="arxiv",
  year={2025}
}

@inproceedings{deng2025mne,
  author       = {Tianchen Deng and
                  Guole Shen and
                  Chen Xun and
                  Shenghai Yuan and
                  Tongxin Jin and
                  Hongming Shen and
                  Yanbo Wang and
                  Jingchuan Wang and
                  Hesheng Wang and
                  Danwei Wang and
                  Weidong Chen
                 },
  title        = {{MNE-SLAM:} Multi-Agent Neural {SLAM} for Mobile Robots},
  booktitle    = {Proceedings of the {IEEE/CVF} Conference on Computer Vision and Pattern Recognition},
  pages        = {1485--1494},
  publisher    = {CVPR},
  year         = {2025} 
}

@inproceedings{song2017sscnet,
  author       = {Shuran Song and
                  Fisher Yu and
                  Andy Zeng and
                  Angel X. Chang and
                  Manolis Savva and
                  Thomas A. Funkhouser
                 },
  title        = {Semantic Scene Completion from a Single Depth Image}, 
  booktitle    = {Proceedings of the {IEEE} Conference on Computer Vision and Pattern Recognition}, 
  pages        = {190--198}, 
  publisher    = {CVPR}, 
  year         = {2017}, 
}

@inproceedings{zhang2019cascaded,
  author       = {Pingping Zhang and
                  Wei Liu and
                  Yinjie Lei and
                  Huchuan Lu and
                  Xiaoyun Yang
                 },
  title        = {Cascaded Context Pyramid for Full-Resolution 3D Semantic Scene Completion},
  booktitle    = {Proceedings of the {IEEE/CVF} International Conference on Computer Vision},
  pages        = {7800--7809},
  publisher    = {{ICCV}},
  year         = {2019}
}

@inproceedings{wang2019forknet,
  author       = {Yida Wang and 
                  David Joseph Tan and 
                  Nassir Navab and 
                  Federico Tombari 
                 }, 
  title        = {ForkNet: Multi-Branch Volumetric Semantic Completion From a Single Depth Image},
  booktitle    = {Proceedings of the {IEEE/CVF} International Conference on Computer Vision},
  pages        = {8607--8616},
  publisher    = {{ICCV}},
  year         = {2019}
}

@inproceedings{li2020anisotropic,
  author       = {Jie Li and
                  Kai Han and
                  Peng Wang and
                  Yu Liu and
                  Xia Yuan
                 },
  title        = {Anisotropic Convolutional Networks for 3D Semantic Scene Completion},
  booktitle    = {Proceedings of the {IEEE/CVF} Conference on Computer Vision and Pattern Recognition},
  pages        = {3348--3356},
  publisher    = {{ICCV}},
  year         = {2020} 
}

@inproceedings{li2019rgbd,
  author       = {Jie Li and
                  Yu Liu and
                  Dong Gong and
                  Qinfeng Shi and
                  Xia Yuan and
                  Chunxia Zhao and
                  Ian D. Reid
                 }, 
  title        = {{RGBD} Based Dimensional Decomposition Residual Network for 3D Semantic
                  Scene Completion},
  booktitle    = {Proceedings of the {IEEE} Conference on Computer Vision and Pattern Recognition},
  pages        = {7693--7702},
  publisher    = {{CVPR}},
  year         = {2019}
}

@inproceedings{wang2023semantic,
  author       = {Fengyun Wang and
                  Dong Zhang and
                  Hanwang Zhang and
                  Jinhui Tang and
                  Qianru Sun
                 },
  title        = {Semantic Scene Completion with Cleaner Self},
  booktitle    = {Proceedings of the {IEEE/CVF} Conference on Computer Vision and Pattern Recognition},
  pages        = {867--877},
  publisher    = {{CVPR}},
  year         = {2023}
}

@inproceedings{roldao2020lmscnet,
  author       = {Luis Rold{\~{a}}o and
                  Raoul de Charette and
                  Anne Verroust{-}Blondet
                 },
  title        = {LMSCNet: Lightweight Multiscale 3D Semantic Completion},
  booktitle    = {8th International Conference on 3D Vision},
  pages        = {111--119},
  publisher    = {{3DV}}, 
  year         = {2020}
}

@inproceedings{yan2021sparse,
  author       = {Xu Yan and
                  Jiantao Gao and
                  Jie Li and
                  Ruimao Zhang and
                  Zhen Li and
                  Rui Huang and
                  Shuguang Cui
                 },
  title        = {Sparse Single Sweep LiDAR Point Cloud Segmentation via Learning Contextual
                  Shape Priors from Scene Completion},
  booktitle    = {Proceedings of the {AAAI} Conference on Artificial Intelligence},
  pages        = {3101--3109},
  publisher    = {{AAAI Press}},
  year         = {2021}
}

@inproceedings{yang2021semantic,
  author       = {Xuemeng Yang and
                  Hao Zou and
                  Xin Kong and
                  Tianxin Huang and
                  Yong Liu and
                  Wanlong Li and
                  Feng Wen and
                  Hongbo Zhang
                 },
  title        = {Semantic Segmentation-assisted Scene Completion for LiDAR Point Clouds},
  booktitle    = {{IEEE/RSJ} International Conference on Intelligent Robots and Systems},
  pages        = {3555--3562},
  publisher    = {{IROS}},
  year         = {2021}
}

@inproceedings{cao2024monoscene,
  author       = {Anh{-}Quan Cao and
                  Raoul de Charette
                 },
  title        = {MonoScene: Monocular 3D Semantic Scene Completion},
  booktitle    = {Proceedings of the {IEEE/CVF} Conference on Computer Vision and Pattern Recognition},
  pages        = {3981--3991},
  publisher    = {{CVPR}},
  year         = {2022}
}

@misc{miao2023occdepth,
  title={Occdepth: A depth-aware method for 3d semantic scene completion},
  author={Miao, Ruihang and Liu, Weizhou and Chen, Mingrui and Gong, Zheng and Xu, Weixin and Hu, Chen and Zhou, Shuchang},
  eprint={2302.13540},
  archivePrefix="arxiv",
  year={2023}
}

@inproceedings{li2023voxformer,
  author       = {Yiming Li and
                  Zhiding Yu and
                  Christopher B. Choy and
                  Chaowei Xiao and
                  Jos{\'{e}} M. {\'{A}}lvarez and
                  Sanja Fidler and
                  Chen Feng and
                  Anima Anandkumar
                 },
  title        = {VoxFormer: Sparse Voxel Transformer for Camera-Based 3D Semantic Scene
                  Completion},
  booktitle    = {Proceedings of the {IEEE/CVF} Conference on Computer Vision and Pattern Recognition},
  pages        = {9087--9098},
  publisher    = {{CVPR}},
  year         = {2023}
}

@article{mei2024camera,
  author       = {Jianbiao Mei and
                  Yu Yang and
                  Mengmeng Wang and
                  Junyu Zhu and
                  Jongwon Ra and
                  Yukai Ma and
                  Laijian Li and
                  Yong Liu
                 },
  title        = {Camera-Based 3D Semantic Scene Completion With Sparse Guidance Network},
  journal      = {{IEEE} Transactions on Image Processing},
  volume       = {33},
  pages        = {5468--5481},
  year         = {2024}
}

@inproceedings{zhu2024CGFormer,
  author = {Yu, Zhu and Zhang, Runmin and Ying, Jiacheng and Yu, Junchen and Hu, Xiaohai and Luo, Lun and Cao, Si-Yuan and Shen, Hui-liang},
  title = {Context and Geometry Aware Voxel Transformer for Semantic Scene Completion},
  booktitle = {Advances in Neural Information Processing Systems},
  volume = {37},
  pages = {1531--1555},
  publisher={{NeurIPS}},
  year = {2024}
}

@inproceedings{jiang2024symphonize,
  author       = {Haoyi Jiang and
                  Tianheng Cheng and
                  Naiyu Gao and
                  Haoyang Zhang and
                  Tianwei Lin and
                  Wenyu Liu and
                  Xinggang Wang
                 },
  title        = {Symphonize 3D Semantic Scene Completion with Contextual Instance Queries},
  booktitle    = {Proceedings of the {IEEE/CVF} Conference on Computer Vision and Pattern Recognition},
  pages        = {20258--20267},
  publisher    = {{CVPR}},
  year         = {2024}
}

@inproceedings{wu2024embodiedocc,
  author       = {Yuqi Wu and
                  Wenzhao Zheng and
                  Sicheng Zuo and
                  Yuanhui Huang and
                  Jie Zhou and
                  Jiwen Lu
                 },
  title        = {EmbodiedOcc: Embodied 3D Occupancy Prediction for Vision-based Online Scene Understanding},
  booktitle    = {Proceedings of the {IEEE/CVF} International Conference on Computer Vision},
  publisher    = {{ICCV}}, 
  year         = {2025},
}

@inproceedings{wang2025embodiedoccplusplus,
  title= {EmbodiedOcc\texttt{++}: Boosting Embodied 3D Occupancy Prediction with Plane Regularization and Uncertainty Sampler},
  author= {Wang, Hao and Wei, Xiaobao and Zhang, Xiaoan and Li, Jianing and Bai, Chengyu and Li, Ying and Lu, Ming and Zheng, Wenzhao and Zhang, Shanghang},
  booktitle= {Proceedings of the 33rd ACM International Conference on Multimedia},
  publisher= {{MM}}, 
  year= {2025},  
}

@misc{zhang2025roboocc,
  title={Roboocc: Enhancing the geometric and semantic scene understanding for robots},
  author={Zhang, Zhang and Zhang, Qiang and Cui, Wei and Shi, Shuai and Guo, Yijie and Han, Gang and Zhao, Wen and Ren, Hengle and Xu, Renjing and Tang, Jian},
  eprint={2504.14604},
  archivePrefix="arxiv",
  year={2025}
}

@inproceedings{yu2024monocular,
  author={Yu, Hongxiao and Wang, Yuqi and Chen, Yuntao and Zhang, Zhaoxiang},
  title={Monocular occupancy prediction for scalable indoor scenes},
  booktitle={Proceedings of the European Conference on Computer Vision},
  volume= {15088}, 
  pages={38--54},
  publisher={ECCV}, 
  year={2024}, 
}

@inproceedings{ronneberger2015u,
  author       = {Olaf Ronneberger and
                  Philipp Fischer and
                  Thomas Brox
                 },
  title        = {U-Net: Convolutional Networks for Biomedical Image Segmentation},
  booktitle    = {Medical Image Computing and Computer-Assisted Intervention},
  volume       = {9351},
  pages        = {234--241},
  publisher    = {MICCAI},
  year         = {2015}
}

@inproceedings{tan2019efficientnet,
  author       = {Mingxing Tan and
                  Quoc V. Le
                 }, 
  title        = {EfficientNet: Rethinking Model Scaling for Convolutional Neural Networks},
  booktitle    = {Proceedings of the 36th International Conference on Machine Learning},
  volume       = {97},
  pages        = {6105--6114},
  publisher    = {{PMLR}}, 
  year         = {2019}
}

@inproceedings{lin2017feature,
  author       = {Tsung{-}Yi Lin and
                  Piotr Doll{\'{a}}r and 
                  Ross B. Girshick and
                  Kaiming He and
                  Bharath Hariharan and
                  Serge J. Belongie
                 }, 
  title        = {Feature Pyramid Networks for Object Detection},
  booktitle    = {Proceedings of the {IEEE} Conference on Computer Vision and Pattern Recognition},
  pages        = {936--944},
  publisher    = {{CVPR}},
  year         = {2017}
}

@inproceedings{yang2024depth,
  author={Yang, Lihe and Kang, Bingyi and Huang, Zilong and Zhao, Zhen and Xu, Xiaogang and Feng, Jiashi and Zhao, Hengshuang},
  title={Depth anything v2},
  booktitle={Advances in Neural Information Processing Systems},
  volume={37},
  pages={21875--21911},
  publisher={{NeurIPS}},
  year={2024}, 
}

@inproceedings{zhu2020deformable,
  author       = {Xizhou Zhu and
                  Weijie Su and
                  Lewei Lu and
                  Bin Li and
                  Xiaogang Wang and
                  Jifeng Dai
                 },
  title        = {Deformable {DETR:} Deformable Transformers for End-to-End Object Detection},
  booktitle    = {Proceedings of the nineth International Conference on Learning Representations},
  publisher    = {{ICLR}},
  year         = {2021}
}

@inproceedings{ashish2017transformer,
  author       = {Ashish Vaswani and
                  Noam Shazeer and
                  Niki Parmar and
                  Jakob Uszkoreit and
                  Llion Jones and
                  Aidan N. Gomez and
                  Lukasz Kaiser and
                  Illia Polosukhin
                 },
  title        = {Attention is All you Need},
  booktitle    = {Advances in Neural Information Processing Systems},
  volume    = {30},
  pages        = {5998--6008},
  publisher    = {{NeurIPS}},
  year         = {2017}
}

@inproceedings{ma2020auto,
  author={Benteng Ma and
          Jing Zhang and
          Yong Xia and
          Dacheng Tao
        },
  title={Auto learning attention},
  booktitle={Advances in neural information processing systems},
  volume={33}, 
  pages={1488--1500},
  publisher={{NeurIPS}},
  year={2020}
}

@inproceedings{jia2025gated,
  author       = {Xiaogang Jia and
                  Songlei Jian and
                  Yusong Tan and
                  Yonggang Che and
                  Wei Chen and
                  Zhengfa Liang
                 },
  title        = {Gated Cross-Attention Network for Depth Completion},
  booktitle    = {Proceedings of the {IEEE} International Conference on Acoustics, Speech and Signal Processing},
  pages        = {1--5},
  publisher    = {{ICASSP}},
  year         = {2025}
}

@inproceedings{wang2025vggt,
  author       = {Jianyuan Wang and
                  Minghao Chen and
                  Nikita Karaev and
                  Andrea Vedaldi and
                  Christian Rupprecht and
                  David Novotn{\'{y}}
                 },
  title        = {{VGGT:} Visual Geometry Grounded Transformer},
  booktitle    = {Proceedings of the {IEEE/CVF} Conference on Computer Vision and Pattern Recognition},
  pages        = {5294--5306},
  publisher    = {{CVPR}},
  year         = {2025}
}

@inproceedings{laina2016deeper,
  author       = {Iro Laina and
                  Christian Rupprecht and
                  Vasileios Belagiannis and
                  Federico Tombari and
                  Nassir Navab
                 },
  title        = {Deeper Depth Prediction with Fully Convolutional Residual Networks},
  booktitle    = {14th International Conference on 3D Vision},
  pages        = {239--248},
  publisher    = {{3DV}},
  year         = {2016}
}

@inproceedings{loshchilov2017decoupled,
  author       = {Ilya Loshchilov and
                  Frank Hutter
                 },
  title        = {Decoupled Weight Decay Regularization},
  booktitle    = {seventh International Conference on Learning Representations},
  publisher    = {{ICLR}},
  year         = {2019}
}

@article{hu2024metric3d,
  author       = {Mu Hu and
                  Wei Yin and
                  Chi Zhang and
                  Zhipeng Cai and
                  Xiaoxiao Long and
                  Hao Chen and
                  Kaixuan Wang and
                  Gang Yu and
                  Chunhua Shen and
                  Shaojie Shen
                 },
  title        = {Metric3D v2: {A} Versatile Monocular Geometric Foundation Model for
                  Zero-Shot Metric Depth and Surface Normal Estimation},
  journal      = {{IEEE} Trans. Pattern Anal. Mach. Intell.},
  volume       = {46},
  number       = {12},
  pages        = {10579--10596},
  publisher    = {TPAMI},
  year         = {2024}
}

@inproceedings{wang2024embodiedscan,
  author       = {Tai Wang and
                  Xiaohan Mao and
                  Chenming Zhu and
                  Runsen Xu and
                  Ruiyuan Lyu and
                  Peisen Li and
                  Xiao Chen and
                  Wenwei Zhang and
                  Kai Chen and
                  Tianfan Xue and
                  Xihui Liu and
                  Cewu Lu and
                  Dahua Lin and
                  Jiangmiao Pang
                 },
  title        = {EmbodiedScan: {A} Holistic Multi-Modal 3D Perception Suite Towards
                  Embodied {AI}},
  booktitle    = {Proceedings of the {IEEE/CVF} Conference on Computer Vision and Pattern Recognition},
  pages        = {19757--19767},
  publisher    = {{CVPR}},
  year         = {2024} 
}

@misc{halacheva2025gaussianvlm,
  title={GaussianVLM: Scene-centric 3D Vision-Language Models using Language-aligned Gaussian Splats for Embodied Reasoning and Beyond},
  author={Halacheva, Anna-Maria and Zaech, Jan-Nico and Wang, Xi and Paudel, Danda Pani and Van Gool, Luc},
  eprint={2507.00886},
  archivePrefix="arxiv",
  year={2025}
}

@article{li2025flagdroneacing,
author = {Li, Ruocheng and Lyu, Jingshuo and Wang, Aobo and Yu, Rui and Wu, Delong and Xin, Bin},
title = {FLAGDroneRacing: An Autonomous Drone Racing System},
journal = {Unmanned Systems},
volume = {12}, 
number = {06}, 
pages = {985-1000},
year = {2024}
}

@article{yi2025embodiedgame,
author = {Yi, Peng and Lei, Jinlong and Hong, Yiguang and Chen, Jie},
title = {Embodied Intelligent Game: Models and Algorithms for Autonomous Interactions Among Heterogeneous Agents},
journal = {Unmanned Systems},
volume = {13},
number = {05},
pages = {1365-1394},
year = {2025}
}

@article{fusic2024improvedrrt,
author = {Fusic, S. Julius and Sitharthan, R.},
title = {Improved RRT* Algorithm-Based Path Planning for Unmanned Aerial Vehicle in a 3D Metropolitan Environment},
journal = {Unmanned Systems},
volume = {12},
number = {05},
pages = {859-875},
year = {2024}
}

@article{zheng2025robustefficient,
author = {Zheng, Zhewen and Cao, Wenjing and Kubota, Yuya and Nakano, Yoshihisa and Gao, Shuang and Suzuki, Takashi},
title = {Robust and Energy-Efficient Torque Vectoring for a Four in-Wheel Motor Electric Vehicle Based on Sliding Mode and Model Predictive Control},
journal = {Unmanned Systems},
volume = {13},
number = {06},
pages = {1699-1712},
year = {2025}
}

@article{zammit2023realtimeuav,
author = {Zammit, Christian and van Kampen, Erik-Jan},
title = {Real-time 3D UAV Path Planning in Dynamic Environments with Uncertainty},
journal = {Unmanned Systems},
volume = {11},
number = {03},
pages = {203-219},
year = {2023}
}
